\documentclass{article}
\usepackage{iclr2027_conference,times}
\usepackage[T1]{fontenc}
\usepackage{graphicx}
\usepackage{hyperref}
\usepackage{url}
\usepackage{amsmath,amssymb}
\usepackage{fix-cm}
\usepackage{tikz}
\usepackage{booktabs,array,multirow}
\usepackage{colortbl,xcolor}
\usepackage{caption}
\usepackage{adjustbox}
\usepackage{placeins}
\usepackage{pifont}
\usetikzlibrary{arrows.meta,positioning}
\pdfmapline{+exp-pala PalatinoLinotype "T1Encoding ReEncodeFont" <fonts/palatino-t1.enc <fonts/exp-pala.ttf}
\pdfmapline{+exp-palab PalatinoLinotype-Bold "T1Encoding ReEncodeFont" <fonts/palatino-t1.enc <fonts/exp-palab.ttf}
\pdfmapline{+exp-palai PalatinoLinotype-Italic "T1Encoding ReEncodeFont" <fonts/palatino-t1.enc <fonts/exp-palai.ttf}
\DeclareFontFamily{T1}{exppl}{}
\DeclareFontShape{T1}{exppl}{m}{n}{<-> fonts/exp-pala}{}
\DeclareFontShape{T1}{exppl}{b}{n}{<-> fonts/exp-palab}{}
\DeclareFontShape{T1}{exppl}{bx}{n}{<-> ssub * exppl/b/n}{}
\DeclareFontShape{T1}{exppl}{m}{it}{<-> fonts/exp-palai}{}
\newcommand{\expfont}{\fontfamily{exppl}\selectfont}
\definecolor{expink}{HTML}{243B47}
\definecolor{expteal}{HTML}{258C9A}
\definecolor{exptint}{HTML}{EFF7F7}
\definecolor{expbaseline}{HTML}{E8E4DD}
\definecolor{expmark}{HTML}{D2622A}
\newcommand{\expbest}[1]{\textcolor{expteal}{\textbf{#1}}}
\newcommand{\expsecond}[1]{\underline{#1}}
\newcommand{\expbase}[1]{\textbf{#1}}
\newcommand{\expfloatstyle}{\expfont\fontsize{8.5}{10}\selectfont}

\iclrfinalcopy
\title{DreamHMR: From Video Generation to Human Motion Modeling via Structural Focusing}
\author{
Zhengjie Xu\\
Beihang University \\
\texttt{25371348@buaa.edu.cn}
\And
Ruihao Gong\thanks{Corresponding author.} \\
Beihang University \\
\texttt{gongruihao@buaa.edu.cn}
\And
Xianglong Liu \\
Beihang University \\
\texttt{xlliu@buaa.edu.cn}
}
\hypersetup{hidelinks,pdftitle={DreamHMR},pdfauthor={Xu Zhengjie, Ruihao Gong, Xianglong Liu}}

\begin{document}
\maketitle

\begin{abstract}

Human mesh recovery (HMR) provides a geometric foundation for understanding human motion, but accurately reconstructing both body motion and fine hand articulation from videos remains challenging, especially in real-world scenes where occlusion and partial visibility are common.
Video generation pretraining offers rich spatial and temporal priors for this task, but its focus on visual synthesis leaves a gap between these representations and explicit human geometry.
We present DreamHMR, a general framework that adapts video generation priors for complete body-and-hand reconstruction.
It follows a global-to-local strategy, first grounding video representations in human geometry and then refining body and hand detail with reliable local evidence.
Global Body Representation Transfer trains the video backbone to predict complete human renderings, turning its native prediction task into a means of learning surface correspondence and posed geometry.
Local Joint Feature Enhancement combines contextual estimates with reliable local evidence to recover fine body and hand detail, while limiting interference from occluders and background.
Experiments on five datasets with four video backbones demonstrate strong reconstruction performance across backbone families.
Under a unified evaluation protocol, our best model outperforms the best existing methods on all reported metrics of 3DPW, EMDB-1, and RICH, reducing body MPJPE by 4.2\%, 4.7\%, and 6.1\%, respectively.
In more challenging occlusion settings on held-out BEDLAM and EgoBody sequences, DreamHMR reduces overall body and hand MPJPE by 7.0\% and 10.0\%, respectively.
Our code will be made publicly available.

\end{abstract}

\section{Introduction}
\label{sec:introduction}

Human mesh recovery (HMR) reconstructs the 3D human body from visual observations, providing a geometric foundation for understanding and modeling human motion.
Its estimates support a wide range of downstream applications, from structured motion analysis for action recognition~\citep{goel2023humans} to motion references for humanoid robot teleoperation and imitation learning~\citep{he2024h2o,fu2025humanplus,zhao2025bilevel,luo2025sonic}.
The field has advanced rapidly, with progress in monocular parametric body estimation~\citep{kanazawa2018endtoend,kolotouros2019spin}, temporal modeling from video~\citep{kocabas2020vibe}, and large-scale learning from diverse imagery~\citep{goel2023humans}.
Despite these advances, recovering complete body and hand motion from real-world videos remains challenging, as occlusion and partial visibility leave incomplete and uneven visual evidence across body parts and frames.

Recovering motion from such observations requires both inferring missing structure and identifying reliable local evidence.
Information from visible body parts and neighboring frames can help constrain unobserved body and hand configurations.
However, local features sampled near a joint may correspond to an occluder or the background rather than the target person, so incorporating them indiscriminately can introduce misleading cues.
A reconstruction model therefore needs to combine contextual inference of the complete body with selective use of local observations for detailed body and hand recovery.

Video generation pretraining offers a promising starting point for learning such contextual representations~\citep{wan2025wan}.
By learning to synthesize videos, these models capture spatial appearance and temporal dependencies within a shared representation, offering potentially useful cues for reasoning about body and hand motion under partial observations.
However, their training objective targets visual synthesis rather than explicit human pose and surface geometry.
Transferring these representations to human reconstruction therefore requires adapting them to the underlying structure of the human body.

To this end, we present DreamHMR, a framework that adapts video generation representations for complete body-and-hand reconstruction through geometric supervision and evidence-gated local refinement.
First, \textit{Global Body Representation Transfer} trains the video backbone to predict renderings of complete SMPL-X bodies~\citep{pavlakos2019smplx}, turning its video prediction objective into a proxy for learning human structure.
Canonical surface colors encode correspondence across poses, while shading conveys pose-dependent geometry, providing supervision for body regions even when they are absent from the observations.
Second, \textit{Local Joint Feature Enhancement} refines contextual body and hand estimates using features sampled around predicted joints.
Supervised evidence scores predict target-person support at these locations and weight the local updates, limiting the influence of features drawn from occluders or background.

Experiments spanning five datasets and four video backbones demonstrate strong reconstruction performance, with gains extending across backbone families.
On standard benchmarks, our Wan-5B model outperforms the best existing methods across all reported metrics under a unified evaluation protocol, reducing body MPJPE by 4.2\%, 4.7\%, and 6.1\% on 3DPW, EMDB-1, and RICH, respectively.
In challenging occlusion settings on held-out BEDLAM and EgoBody sequences, it reduces overall body and hand MPJPE by 7.0\% and 10.0\%, respectively.

DreamHMR makes the following contributions:
\begin{itemize}
    \item We bring pretrained video generation priors to complete body-and-hand reconstruction through \textit{Global Body Representation Transfer}, offering a new approach to recovering human motion from occluded and incomplete observations.
    \item We introduce \textit{Local Joint Feature Enhancement} to recover fine body and hand detail from reliable local evidence, complementing global reconstruction within a general framework applicable across different video backbones.
    \item Extensive experiments on five datasets and four video backbones demonstrate that DreamHMR outperforms the best existing methods on standard HMR benchmarks, with body and hand MPJPE reduced by 7.0\% and 10.0\%, respectively, in challenging occlusion settings.
\end{itemize}

\section{Related Work}
\label{sec:related-work}

\paragraph{Human reconstruction from images and videos.}
\label{sec:rw-hmr}
Human mesh recovery estimates parametric bodies from images~\citep{kanazawa2018endtoend,kolotouros2019spin}, with video models adding temporal context~\citep{kocabas2020vibe,goel2023humans}. PromptHMR-Vid combines full-image features with a box-prompted temporal transformer~\citep{wang2025prompthmr}; GENMO learns a motion distribution for generation and estimation~\citep{li2025genmo}. Whole-body methods recover articulated bodies and hands~\citep{pavlakos2019smplx,lin2023osx,li2023hybrik,baradel2024multihmr,moon2026hand4wholepp}, scale training data and model capacity~\citep{cai2023smplerx,yin2026smplestx,yang2026sam3dbody}; dedicated models recover detailed hand meshes~\citep{pavlakos2024hamer,potamias2025wilor}. DreamHMR uses a shared video generation representation for body-and-hand recovery.

\paragraph{Generative representations for reconstruction.}
\label{sec:rw-generative}
Image diffusion features support correspondence estimation~\citep{tang2023dift,stracke2025cleandift}, while video diffusion features support motion and geometry tasks~\citep{velez2025imagetovideo,nam2025difftrack,huang2026howmuch3d}. DPMesh conditions a text-to-image diffusion backbone on 2D observations for occluded mesh recovery~\citep{zhu2024dpmesh}, and ADHMR optimizes diffusion-based mesh predictions with preference data~\citep{shen2025adhmr}. DreamHMR adapts video generation features with complete human renderings to encode surface correspondence and posed geometry for body-and-hand recovery.

\paragraph{Reconstruction under occlusion.}
\label{sec:rw-limited}
Occluded HMR addresses body reconstruction with incomplete observations~\citep{khirodkar2022occluded}. PARE aggregates neighboring regions with part-guided attention~\citep{kocabas2021pare}, and Divide and Fuse merges part meshes from partially visible images~\citep{luan2024divideandfuse}. Human and joint tokens structure decoding~\citep{sun2024aios}, while local image-aligned features refine hands~\citep{potamias2025wilor}. Nearby features may describe background or another person; DreamHMR learns evidence scores under supervision to weight their updates.

\section{Method}
\label{sec:method}

DreamHMR recovers complete body-and-hands motion through Global Body Representation Transfer, which adapts video features with complete rendered human structure, and Local Joint Feature Enhancement, which refines anatomical states with local image evidence.

\subsection{Overall framework}
\label{sec:task}

Given target-person RGB $I_{1:T}$, binary observation masks $M_{1:T}$, and frozen-tracker boxes $B_{1:T}$, DreamHMR predicts complete SMPL-X motion, body shape, and presence:
\begin{equation}
    \bigl(\widehat{\boldsymbol\phi}_{1:T},\widehat{\boldsymbol\theta}_{1:T},
    \widehat{\mathbf t}_{1:T},\widehat{\boldsymbol\beta},\widehat a_{1:T}\bigr)
    = f(I_{1:T},M_{1:T},B_{1:T}),
    \label{eq:task}
\end{equation}
where $\boldsymbol\phi_t$ is the root orientation, $\boldsymbol\theta_t$ contains the local rotations of the 21 body joints and 15 joints in each hand, and $\mathbf t_t$ is the camera-frame pelvis position. The shape vector $\boldsymbol\beta\in\mathbb R^{10}$ is shared across the window, and $a_t$ denotes the target person's presence in the observed canvas.

\begin{figure}[!t]
    \centering
    \includegraphics[width=\linewidth]{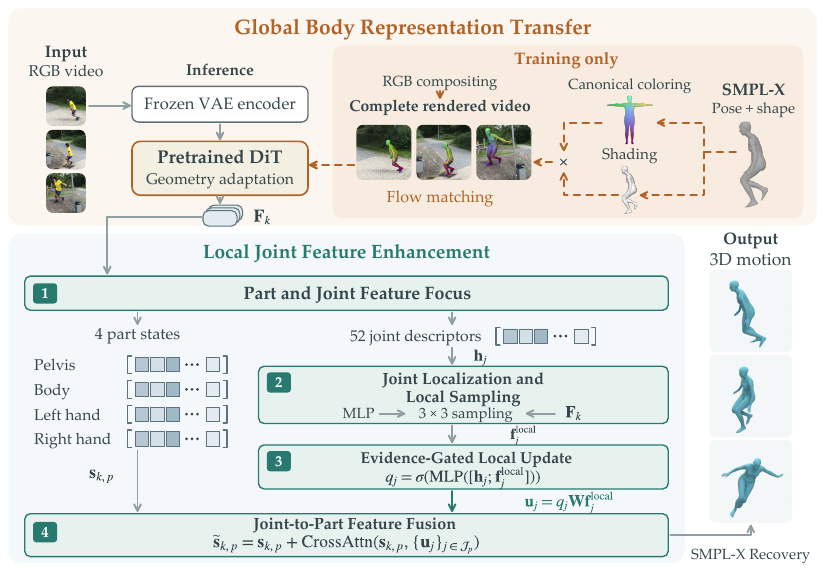}
    \caption{\textbf{DreamHMR architecture.} Global Body Representation Transfer adapts a video DiT with complete SMPL-X renderings. Local Joint Feature Enhancement samples joint-local features, weights their corrections using predicted evidence scores, and fuses them into body-and-hand part states.}
    \label{fig:architecture}
\end{figure}

As shown in Figure~\ref{fig:architecture}, DreamHMR uses complete geometry supervision to adapt the video backbone and locally supported refinement to recover details that are only visible in parts of the image. Global Body Representation Transfer renders a complete shaded human video whose colors encode a canonical surface map and whose shading reflects posed geometry. Local Joint Feature Enhancement then samples features around predicted joints, weights their corrections using predicted evidence scores, and fuses the resulting corrections into four anatomical states.

\subsection{Global Body Representation Transfer}
\label{sec:geometry-transfer}
Global Body Representation Transfer injects complete projected human structure into the video representation over the full image canvas, including regions outside the observation mask.

\paragraph{Rendering complete human structure.}
We construct the target video from the annotated SMPL-X pose and shape. For each frame, we place the posed mesh's pelvis at $\mathbf t_t$ to obtain camera-frame vertices $\mathbf V_t$. The camera projection $\Pi_t$ uses the input calibration and the same resized-and-padded canvas, aligning the rendered mesh with the RGB input.

\begin{figure}[t]
    \centering
    \includegraphics[width=\linewidth]{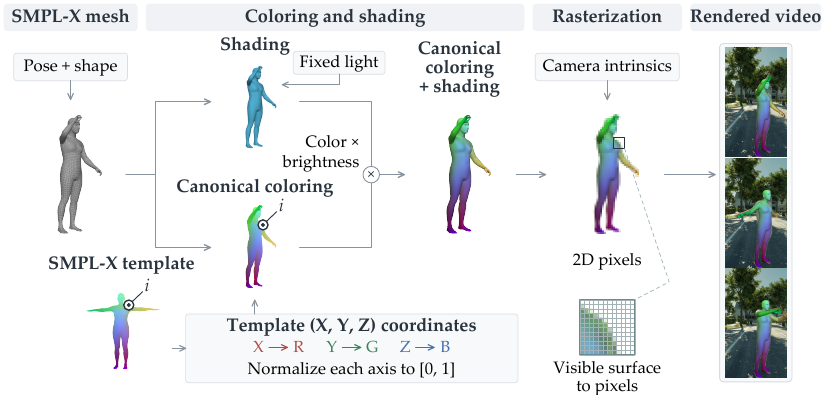}
    \caption{\textbf{From SMPL-X to rendered video.} Fixed template coordinates provide canonical colors; posed normals determine shading. Their product is rasterized and composited with source RGB across frames. Vertex colors persist across poses; a neutral shape and enlarged pixels aid illustration.}
    \label{fig:rendering-process}
\end{figure}

We render a canonical surface map~\citep{kulkarni2019canonical}: every visible pixel is colored by the canonical surface coordinates of its corresponding template point (Figure~\ref{fig:rendering-process}). The image-space encoding of these coordinates follows dense human pose estimation~\citep{guler2018densepose}. Let $\mathbf T_i$ be vertex $i$ of a neutral SMPL-X mesh in its template pose, with per-axis extrema $\mathbf T_{\min}$ and $\mathbf T_{\max}$. We normalize these template coordinates componentwise to obtain the R, G, and B channels:
\begin{equation}
    \mathbf C_i = \frac{\mathbf T_i-\mathbf T_{\min}}
    {\mathbf T_{\max}-\mathbf T_{\min}}\in[0,1]^3.
    \label{eq:geometry-target}
\end{equation}
The canonical surface map is computed once: vertex $i$ retains $\mathbf C_i$ as pose and shape deform the mesh. The rendered color therefore identifies the same body surface location across poses, people, and frames, supplying dense correspondence supervision. This correspondence is defined on the template surface rather than by a point's image location. As the body moves, its projected location can change while the underlying canonical color assigned to that surface point remains the same.

Canonical coloring and shading define the appearance of the same posed mesh: template coordinates supply the surface colors, while current surface normals and a fixed light determine brightness. Their product combines stable surface correspondence with pose-dependent geometry. Rasterization projects the triangles onto the image, resolves visibility, and evaluates this appearance at covered pixels. Compositing with the source RGB yields the three-channel target $I_t^{\rm rend}$; repeating the process across frames produces the supervision video. Rendering covers the full canvas independently of $M_t$, retaining projected human pixels outside the observed region; Appendix~\ref{app:rendering} details the renderer.

\paragraph{Learning from the rendered video.}
A frozen VAE encoder $E$ maps the rendered video to the clean target latent $\mathbf z_0=E(I_{1:T}^{\rm rend})$ and input RGB to $\mathbf z_I=E(I_{1:T})$. The condition $\mathcal C$ comprises the RGB latent $\mathbf z_I$, observation masks, box tracks, and a fixed text context. Following the linear flow-matching path used by Wan~\citep{wan2025wan}, we interpolate between the target latent and Gaussian noise at noise level $\tau\in[0,1]$:
\begin{equation}
    \mathbf z_\tau=(1-\tau)\mathbf z_0+\tau\boldsymbol\epsilon,
    \qquad \boldsymbol\epsilon\sim\mathcal N(\mathbf 0,\mathbf I),
    \label{eq:geometry-flow}
\end{equation}
with $\tau=0$ denoting the clean target latent. The target velocity is $\partial\mathbf z_\tau/\partial\tau=\boldsymbol\epsilon-\mathbf z_0$. We train the DiT velocity predictor $v(\mathbf z_\tau,\tau;\mathcal C)$ with the conditional flow-matching objective
\begin{equation}
    \mathcal L_{\rm FM}=\mathbb E_{\mathbf z_0,\mathcal C,\tau,\boldsymbol\epsilon}\!\left[
    \left\|v(\mathbf z_\tau,\tau;\mathcal C)
    -(\boldsymbol\epsilon-\mathbf z_0)\right\|_2^2\right].
    \label{eq:geometry-loss}
\end{equation}
The expectation is over paired target latents and conditions, noise levels, and Gaussian noise; in training, the squared error is normalized by the number of latent elements. The VAE and text encoder remain frozen. This stage adapts the DiT to predict complete projected human structure and provides the features used by the structured decoder.

\subsection{Local Joint Feature Enhancement}
\label{sec:local-enhancement}

The structured decoder combines global anatomical states with local joint descriptors: states summarize complete body-part context, while descriptors locate joints and retrieve nearby image evidence.

\paragraph{Part and Joint Feature Focus.}
Let $\mathbf F_k$ be the adapted DiT feature map at latent frame $k$, projected to the decoder dimension $d=512$, with $K=21$ latent frames.
Four learned part queries and 52 learned joint queries extract part and joint information from shared features through cross-attention:
\begin{equation}
    \mathbf s_{k,p}=\operatorname{CrossAttn}_{\mathrm{part}}(\mathbf F_k),
    \qquad
    \mathbf h_{k,j}=\operatorname{CrossAttn}_{\mathrm{joint}}(\mathbf F_k, \mathbf s_{k,p}).
    \label{eq:structured-readout}
\end{equation}
Here $p\in\{1,\ldots,4\}$ indexes the pelvis, remaining body, left hand, and right hand, while $j\in\{1,\ldots,52\}$ indexes body and hand joints. Part-state vectors $\mathbf s_{k,p}\in\mathbb R^d$ represent the overall context of each anatomical region, while joint descriptors guide localization and local sampling. Assigning joint corrections to their corresponding parts structures the subsequent feature fusion. A box encoder supplies geometry, confidence, and tracking status as tokens injected at the corresponding times.

\paragraph{Joint Localization and Local Sampling.}
Within each latent frame, each descriptor predicts an image location and retrieves a local feature:
\begin{equation}
    \widehat{\mathbf x}_j=\operatorname{MLP}_{\rm loc}(\mathbf h_j),
    \qquad
    \mathbf f_j^{\rm local}=\operatorname{LocalPool}(\mathbf F_k,\widehat{\mathbf x}_j).
    \label{eq:local-sampling}
\end{equation}
LocalPool bilinearly samples a fixed $3\times3$ neighborhood around the predicted location and averages the valid samples. Available ground-truth joints supervise localization.

\paragraph{Evidence-Gated Local Update.}
An evidence head predicts an evidence score for each joint and uses it to weight the local correction:
\begin{equation}
    q_j=\sigma\!\left(\operatorname{MLP}([\mathbf h_j;\mathbf f_j^{\rm local}])\right),
    \qquad \mathbf u_j=q_j\mathbf W\mathbf f_j^{\rm local},
    \label{eq:support-gate}
\end{equation}
where $q_j$ is the evidence score, $\sigma$ is the sigmoid, $[\cdot;\cdot]$ denotes concatenation, $\mathbf W$ projects the local feature, and $\mathbf u_j$ is the gated correction.

\paragraph{Evidence Supervision.}
We supervise the evidence score $q_j$ with a binary evidence label $q_j^{\rm GT}$. The label is 1 when the ground-truth joint projection lies within both the observation region and the target person's visible instance mask, and 0 otherwise. We train the evidence head with
\begin{equation}
    \mathcal L_{\rm support}=\operatorname{BCE}(q_j,q_j^{\rm GT}).
    \label{eq:evidence-supervision}
\end{equation}
The balanced BCE averages the losses within three groups---supported joints inside the observation region, unsupported joints inside it, and joints outside it---and then equally averages the nonempty groups. Supervision uses available mask and projection annotations at the original frames corresponding to latent endpoints, with group averages computed over the training batch. Reconstruction gradients also train $q_j$ through the gated local update.

\paragraph{Joint-to-Part Feature Fusion.}
Gated corrections update the corresponding anatomical state. Let $\mathcal J_p$ contain the joints assigned to part $p$. Cross-attention fuses their corrections into the part state:
\begin{equation}
    \widetilde{\mathbf s}_{k,p}=\mathbf s_{k,p}+
    \operatorname{CrossAttn}
    \!\left(\mathbf s_{k,p},\{\mathbf u_{j}\}_{j\in\mathcal J_p}\right),
    \label{eq:state-fusion}
\end{equation}
where $\mathbf s_{k,p}$ serves as the query and $\{\mathbf u_j\}_{j\in\mathcal J_p}$ serves as keys and values. Prediction heads map the refined states to root orientation, local rotations, pelvis position, shape, and presence. Temporal recovery maps the $K$ latent endpoints to all $T$ frames, while $\boldsymbol\beta$ remains shared across the window.

\subsection{Training and Inference}
\label{sec:training-inference}

We first adapt the DiT with rendered-video flow matching, then jointly train the backbone and structured decoder for parameter recovery using
\begin{equation}
    \mathcal L_{\rm recovery}=\mathcal L_{\rm SMPLX}+\lambda_{\rm support}\mathcal L_{\rm support},
    \label{eq:readout-loss}
\end{equation}
where $\mathcal L_{\rm SMPLX}$ supervises pose, position, shape, joint localization, temporal velocity, and presence. Training mixes full, static, and moving observations; Appendix~\ref{app:implementation} details losses and freezing.

At inference, the adapted DiT processes the VAE-encoded RGB once at $\tau=0$, conditioned on masks, box tracks, and fixed text context. The structured decoder maps its features to SMPL-X parameters and presence, following the same feature path used in parameter-recovery training.

\section{Experiments}
\label{sec:experiments}

We evaluate DreamHMR on standard human mesh recovery benchmarks and complete body-and-hand recovery under occlusion across four video backbones. Ablations examine rendering supervision, local evidence mechanisms, and temporal context. Our code is available in the supplementary material.

\subsection{Experimental Setup}
\label{sec:exp-setup}
We train on BEDLAM~\citep{black2023bedlam}, BEDLAM2~\citep{tesch2025bedlam2}, EgoBody~\citep{zhang2022egobody}, ARCTIC~\citep{fan2023arctic}, InterCap~\citep{huang2022intercap}, and IDEA400~\citep{lin2023motionx}, using full and spatially masked 81-frame windows. Occlusion evaluation and ablations use separate holdouts of 90 BEDLAM and 30 EgoBody sequences each, disjoint from training and validation.

We compare PromptHMR-Vid~\citep{wang2025prompthmr}, GENMO~\citep{li2025genmo}, SMPLest-X~\citep{yin2026smplestx}, 3DB-DINOv3~\citep{yang2026sam3dbody}, and Hand4Whole++~\citep{moon2026hand4wholepp}, and instantiate DreamHMR with Cosmos-2B~\citep{nvidia2025cosmos}, LTX-2B~\citep{hacohen2024ltxvideo}, Wan-1.3B, and Wan-5B~\citep{wan2025wan}. Methods share observed RGB and target tracks; video methods use full windows, while image methods use track crops without temporal smoothing. We report MPJPE, Procrustes-aligned MPJPE (PA-MPJPE), and per-vertex error (PVE) on 3DPW test~\citep{vonMarcard2018}, EMDB-1~\citep{kaufmann2023emdb}, and RICH test~\citep{huang2022rich}. Occlusion results report pelvis-relative body and wrist-relative hand MPJPE. Errors are in millimeters. Appendices~\ref{app:implementation} and~\ref{app:observations} detail backbones, observation construction, alignment, and aggregation.

\subsection{Evaluation on Widely Used HMR Benchmarks}
\label{sec:exp-public}
\begin{table}[!htb]
\centering
\caption{\textbf{HMR benchmark comparison.} Errors in mm; PA denotes PA-MPJPE. Methods share RGB, targets, joint regressors, alignment, and sequence averaging. Blue bold, underlined, and black bold values mark the best, second-best, and best external result per column, respectively.}
\label{tab:public}
\normalfont\normalsize
\setlength{\tabcolsep}{2pt}
\begin{tabular}{lccccccccc}
\toprule
\multirow{2}{*}{Method} & \multicolumn{3}{c}{3DPW} & \multicolumn{3}{c}{EMDB-1} & \multicolumn{3}{c}{RICH} \\
\cmidrule(lr){2-4}\cmidrule(lr){5-7}\cmidrule(l){8-10}
& PA $\downarrow$ & MPJPE $\downarrow$ & PVE $\downarrow$ & PA $\downarrow$ & MPJPE $\downarrow$ & PVE $\downarrow$ & PA $\downarrow$ & MPJPE $\downarrow$ & PVE $\downarrow$ \\
\midrule
PromptHMR-Vid                   & 39.7              & 59.8              & 67.8              & 39.8              & 68.5              & 78.7          & 37.8          & 57.8          & 65.3 \\
GENMO                           & 34.4              & \expbase{54.2}    & 65.5              & 42.8              & 65.5              & 75.3          & 36.4          & 60.2          & 64.9 \\
SMPLest-X    & 33.8              & 54.5              & 64.1              & 41.1              & 63.0              & 81.0          & \expbase{28.9} & \expbase{52.5} & \expbase{58.7} \\
3DB-DINOv3                      & \expbase{33.5}    & 55.1              & \expbase{63.2}    & \expbase{38.5}    & \expbase{61.2}    & \expbase{72.9} & 30.6         & 54.0          & 59.8 \\
Hand4Whole++                    & 37.5              & 57.0              & 69.6              & 45.6              & 70.2              & 81.4          & 35.3          & 62.0          & 67.2 \\
\midrule
DreamHMR / Cosmos-2B & \expsecond{31.9} & 53.6 & 62.9 & \expsecond{37.2} & \expsecond{60.5} & 72.9 & 30.5 & 53.6 & \expsecond{57.4} \\
DreamHMR / LTX-2B & 33.8 & 53.8 & 63.5 & 38.5 & 62.8 & 73.1 & 31.2 & 53.9 & 58.5 \\
DreamHMR / Wan-1.3B & 32.4 & \expsecond{53.0} & \expsecond{62.3} & 37.9 & 60.9 & \expsecond{71.2} & \expsecond{28.0} & \expsecond{51.0} & 58.0 \\
DreamHMR / Wan-5B & \expbest{30.9} & \expbest{51.9} & \expbest{61.9} & \expbest{36.2} & \expbest{58.3} & \expbest{69.8} & \expbest{27.2} & \expbest{49.3} & \expbest{55.2} \\
\bottomrule
\end{tabular}
\end{table}

DreamHMR attains the lowest error on every metric of the widely used human mesh recovery benchmarks. In Table~\ref{tab:public}, Wan-5B improves on every metric: MPJPE by 2.3 mm on 3DPW, 2.9 mm on EMDB-1, and 3.2 mm on RICH, with PVE reductions of 1.3, 3.1, and 3.5 mm. All four DreamHMR backbones outperform the best external 3DPW body MPJPE result.

\begin{figure}[!htb]
\centering
{\expfloatstyle
\setlength{\tabcolsep}{1.3pt}
\begin{tabular}{@{}ccccc@{}}
RGB & Hand4Whole++ & SMPLest-X & DreamHMR & GT \\[2pt]
\begin{tikzpicture}[baseline=(zoom.south)]\node[inner sep=0,anchor=south west] (im) {\includegraphics[width=.194\linewidth]{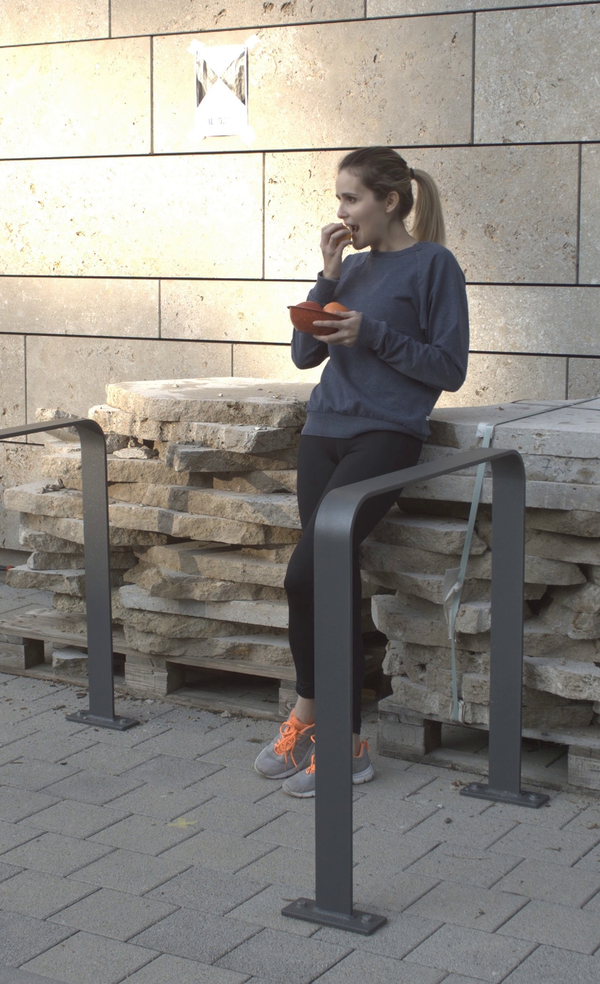}};\node[inner sep=0,anchor=north west] (zoom) at ([yshift=-9pt]im.south west) {\includegraphics[width=.194\linewidth]{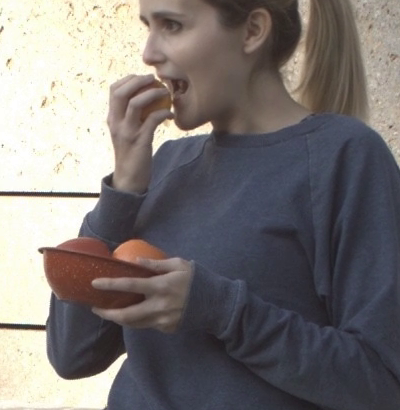}};\begin{scope}[shift={(im.south west)},x=.194\linewidth,y=.31816\linewidth]\draw[expmark,line width=1.15pt] (.448,.610) rectangle (.768,.810);\end{scope}\coordinate (tip) at ([xshift=.021\linewidth,yshift=.19408\linewidth]im.south);\draw[overlay,expmark,line width=1.5pt,-{Latex[length=5pt,width=5.2pt]}] (tip) -- ([xshift=.118\linewidth]zoom.north west);\end{tikzpicture} & \begin{tikzpicture}[baseline=(zoom.south)]\node[inner sep=0,anchor=south west] (im) {\includegraphics[width=.194\linewidth]{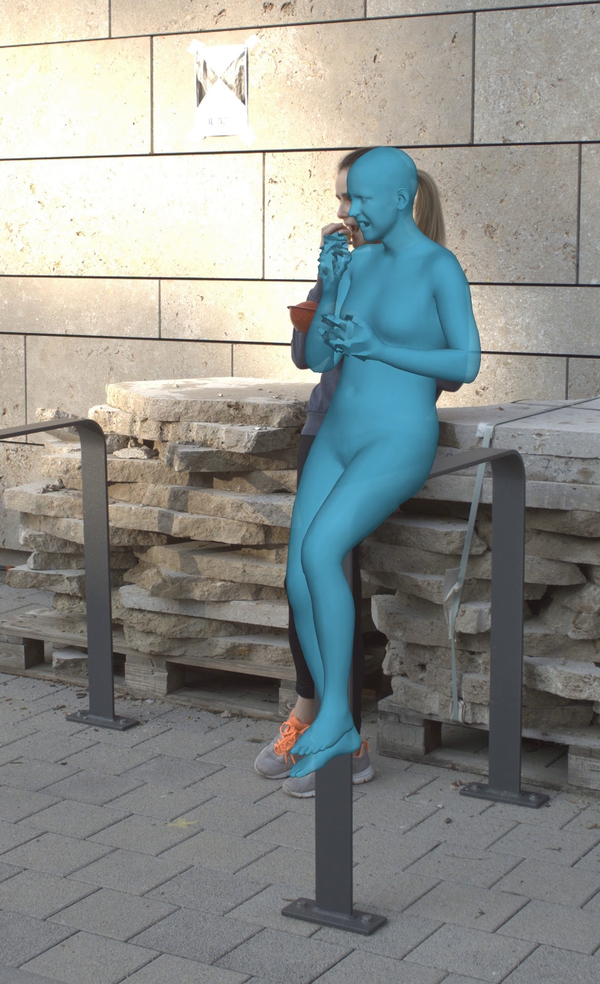}};\node[inner sep=0,anchor=north west] (zoom) at ([yshift=-9pt]im.south west) {\includegraphics[width=.194\linewidth]{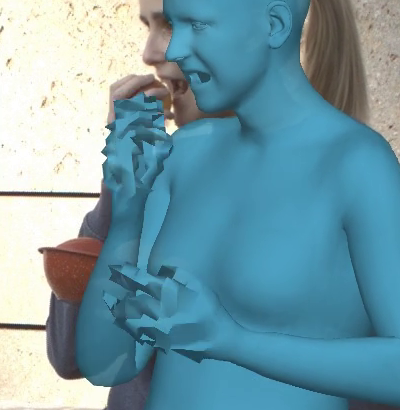}};\end{tikzpicture} & \begin{tikzpicture}[baseline=(zoom.south)]\node[inner sep=0,anchor=south west] (im) {\includegraphics[width=.194\linewidth]{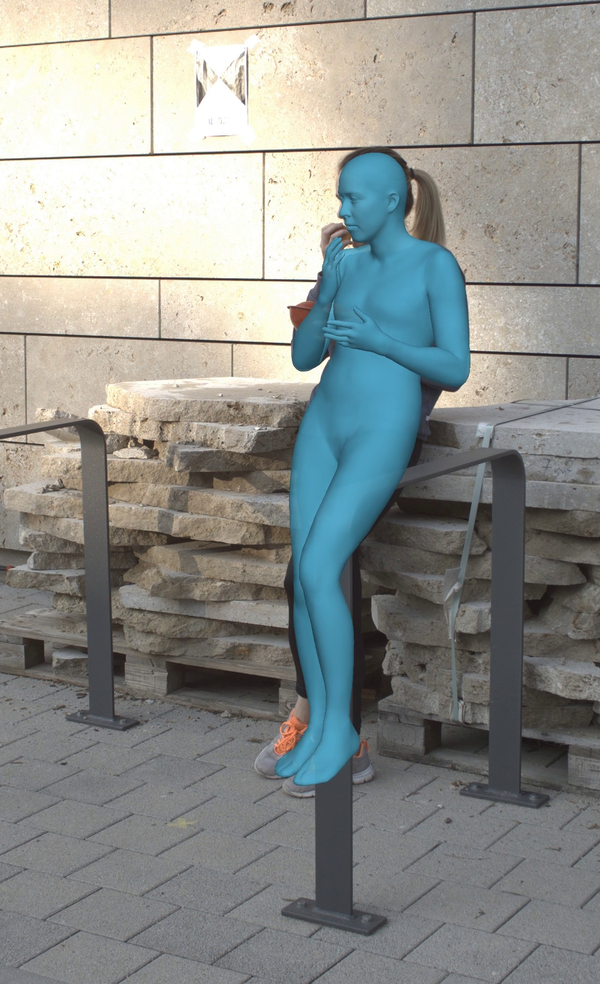}};\node[inner sep=0,anchor=north west] (zoom) at ([yshift=-9pt]im.south west) {\includegraphics[width=.194\linewidth]{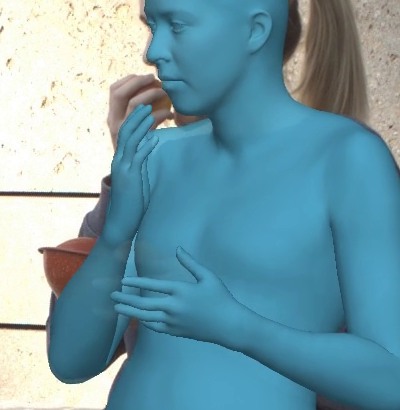}};\end{tikzpicture} & \begin{tikzpicture}[baseline=(zoom.south)]\node[inner sep=0,anchor=south west] (im) {\includegraphics[width=.194\linewidth]{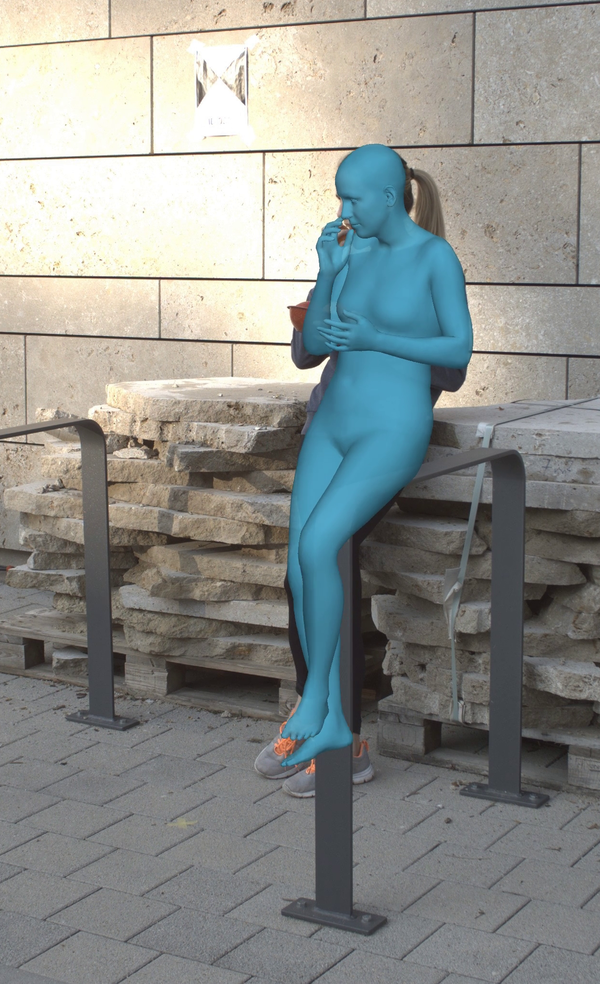}};\node[inner sep=0,anchor=north west] (zoom) at ([yshift=-9pt]im.south west) {\includegraphics[width=.194\linewidth]{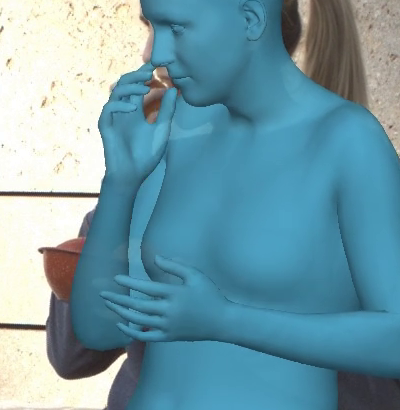}};\end{tikzpicture} & \begin{tikzpicture}[baseline=(zoom.south)]\node[inner sep=0,anchor=south west] (im) {\includegraphics[width=.194\linewidth]{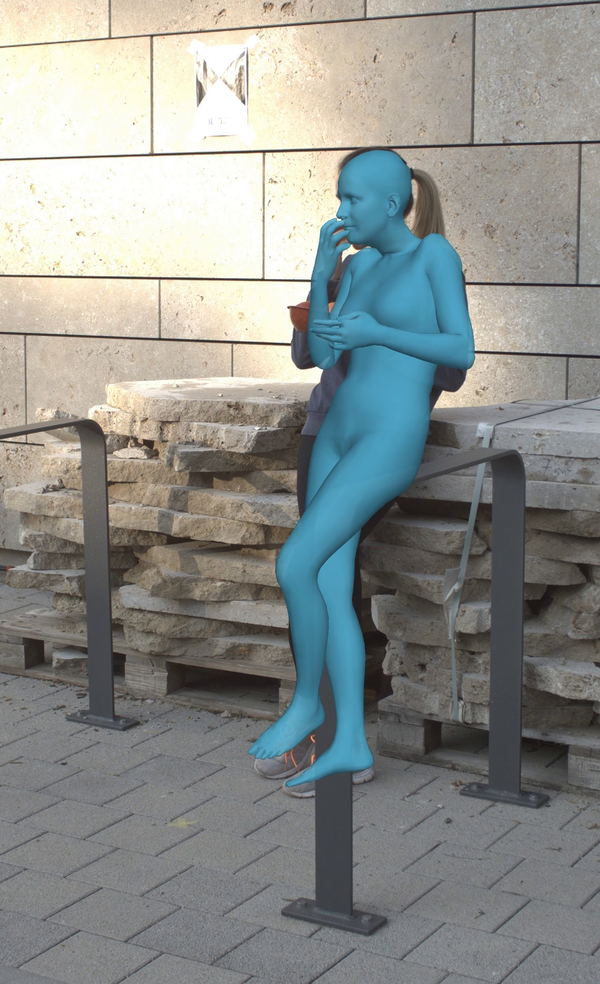}};\node[inner sep=0,anchor=north west] (zoom) at ([yshift=-9pt]im.south west) {\includegraphics[width=.194\linewidth]{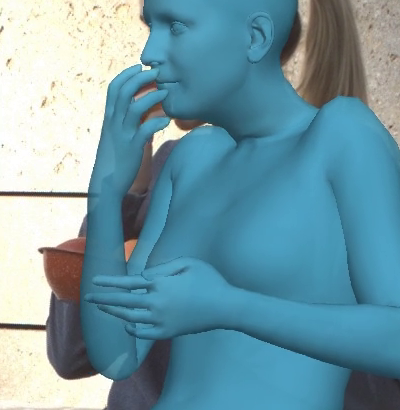}};\end{tikzpicture} \\
\end{tabular}\par
}
\caption{\textbf{Real-scene occlusion and hand detail.} Wan-1.3B on a RICH eating sequence with foreground railings. The box and arrow mark the hand-and-container region enlarged below. Methods share the crop and scale; GT denotes ground-truth meshes.}
\label{fig:public-qualitative}
\end{figure}

Figure~\ref{fig:public-qualitative} shows how this advantage appears in real scenes. In the boxed close-up, SMPLest-X~\citep{yin2026smplestx} lays the fingers flat against the torso, leaving the bowl unheld, while DreamHMR keeps the fingers wrapped around the food and the bowl rim. See Appendix~\ref{app:qualitative} for temporal comparisons.

\subsection{Body-and-Hand Reconstruction under Occlusion}
\label{sec:exp-limited}
\begin{table}[!htb]
\centering
\caption{\textbf{Body-and-hand reconstruction under occlusion.} MPJPE (mm); Overall weights BEDLAM/EgoBody at 75\%/25\% and static/dynamic masks equally. Blue bold and underlined values mark the best and second-best results per column; -- denotes no articulated-hand output.}
\label{tab:limited}
\normalfont\normalsize
\setlength{\tabcolsep}{4pt}
\begin{tabular}{lcccccc}
\toprule
\multirow{2}{*}{Method} & \multicolumn{2}{c}{Overall} & \multicolumn{2}{c}{BEDLAM} & \multicolumn{2}{c}{EgoBody} \\
\cmidrule(lr){2-3}\cmidrule(lr){4-5}\cmidrule(l){6-7}
& Body $\downarrow$ & Hands $\downarrow$ & Body $\downarrow$ & Hands $\downarrow$ & Body $\downarrow$ & Hands $\downarrow$ \\
\midrule
PromptHMR-Vid & 121.2 & -- & 120.6 & -- & 122.8 & -- \\
GENMO & 118.8 & -- & 116.8 & -- & 124.6 & -- \\
SMPLest-X & 115.6 & 61.6 & 113.9 & 60.7 & 120.5 & 64.3 \\
3DB-DINOv3 & 115.2 & 61.2 & 114.4 & 59.4 & 117.6 & 66.6 \\
Hand4Whole++ & 118.3 & 63.1 & 117.0 & 62.4 & 122.2 & 65.2 \\
\midrule
DreamHMR / Cosmos-2B & \expsecond{109.7} & 60.7 & \expsecond{108.2} & 59.8 & 114.2 & 63.4 \\
DreamHMR / LTX-2B & 113.2 & 62.1 & 111.8 & 61.5 & 117.4 & 63.8 \\
DreamHMR / Wan-1.3B & 111.3 & \expsecond{57.5} & 110.5 & \expsecond{55.8} & \expsecond{113.7} & \expsecond{62.6} \\
DreamHMR / Wan-5B & \expbest{107.1} & \expbest{55.1} & \expbest{105.2} & \expbest{53.4} & \expbest{112.6} & \expbest{60.0} \\
\bottomrule
\end{tabular}
\end{table}

DreamHMR improves complete body-and-hand reconstruction under occlusion across both evaluation sources. In Table~\ref{tab:limited}, Wan-5B reaches 107.1/55.1 mm body/hand MPJPE, improving on the strongest external baseline, 3DB-DINOv3, by 8.1/6.1 mm. It also achieves the lowest error in every source-specific column, including EgoBody's real interaction videos.

The gains extend across video backbone families: all four improve on the strongest external overall body result, with Cosmos-2B ranking second, while both Wan models provide the strongest joint body-and-hand results. We use Wan-1.3B for the design analyses below.

\subsection{Ablation Studies}
\label{sec:exp-ablations}
We evaluate Wan-1.3B on the independent ablation holdout, reporting full-window body/hand MPJPE for Full (without additional masking) and Occlusion inputs. Rendering comparisons vary the first-stage target while fixing complete projected coverage and measure body PVE on shared SMPL-X vertices with pelvis alignment, no Procrustes fitting, and common GT-valid frames and vertices. Decoder and gating variants share adapted backbone initialization and reconstruction settings.

\begin{table}[!htb]
\centering
\caption{\textbf{Rendering supervision.} Wan-1.3B errors (mm) over all 81 frames on the ablation holdout. Geometry targets share complete projected coverage. Body/Hands report MPJPE; body PVE uses shared SMPL-X vertices and pelvis alignment. RGB supervision is the reference.}
\label{tab:rendering-ablation}
\normalfont\normalsize
\setlength{\tabcolsep}{4pt}
\begin{tabular}{lcccccc}
\toprule
\multirow{2}{*}{Supervision target} & \multicolumn{3}{c}{Full} & \multicolumn{3}{c}{Occlusion} \\
\cmidrule(lr){2-4}\cmidrule(l){5-7}
& Body $\downarrow$ & Hands $\downarrow$ & PVE $\downarrow$ & Body $\downarrow$ & Hands $\downarrow$ & PVE $\downarrow$ \\
\midrule
RGB & 62.3 & 39.2 & 73.4 & 120.0 & 68.4 & 140.6 \\
Shading & 55.9 & 32.4 & 64.2 & \expsecond{109.3} & 64.8 & \expsecond{126.7} \\
Canonical coloring & 56.1 & 32.8 & 66.8 & 111.6 & \expsecond{60.7} & 129.1 \\
Shading + skeleton & \expsecond{54.7} & \expsecond{31.3} & \expsecond{62.5} & 110.5 & 64.2 & 127.4 \\
Shading + canonical coloring & \expbest{53.8} & \expbest{30.7} & \expbest{61.3} & \expbest{106.3} & \expbest{57.8} & \expbest{121.8} \\
\bottomrule
\end{tabular}
\end{table}

\begin{figure}[!htb]
\centering
{\expfloatstyle\fontsize{8}{9.5}\selectfont
\setlength{\tabcolsep}{1.4pt}
\begin{tabular}{@{}ccccc@{}}
\includegraphics[width=.193\linewidth]{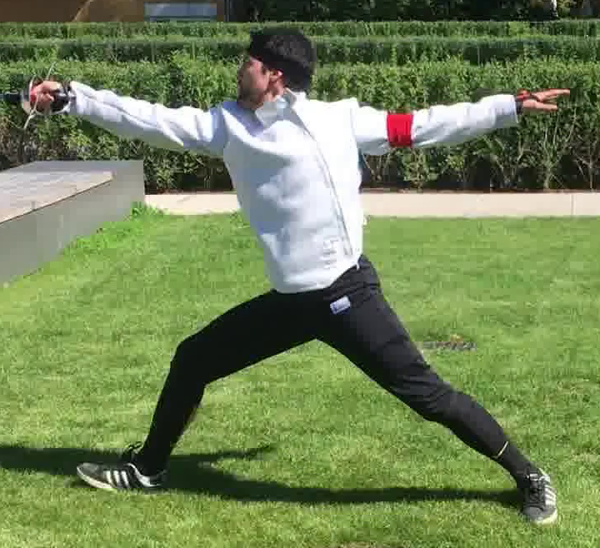} & \includegraphics[width=.193\linewidth]{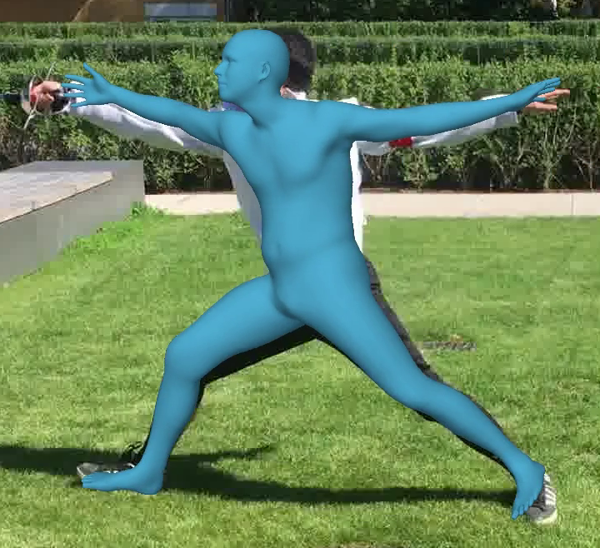} & \includegraphics[width=.193\linewidth]{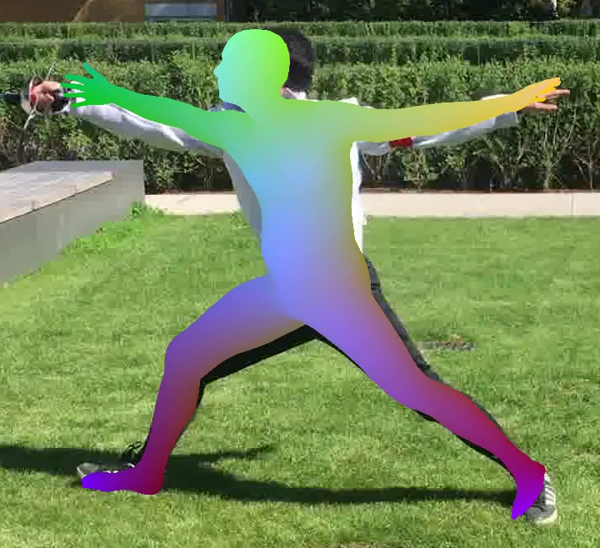} & \includegraphics[width=.193\linewidth]{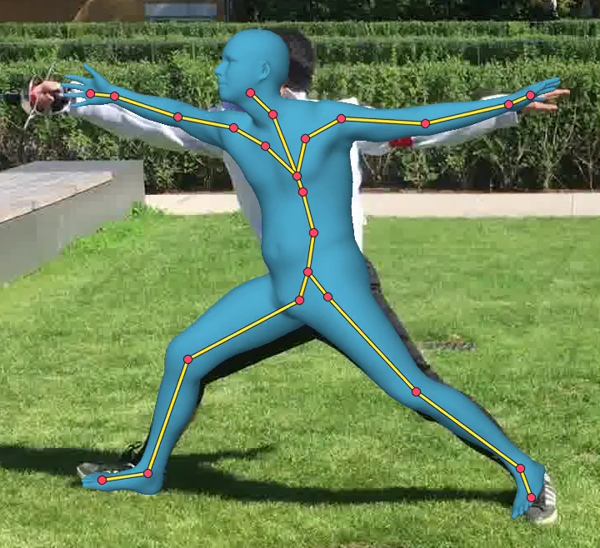} & \includegraphics[width=.193\linewidth]{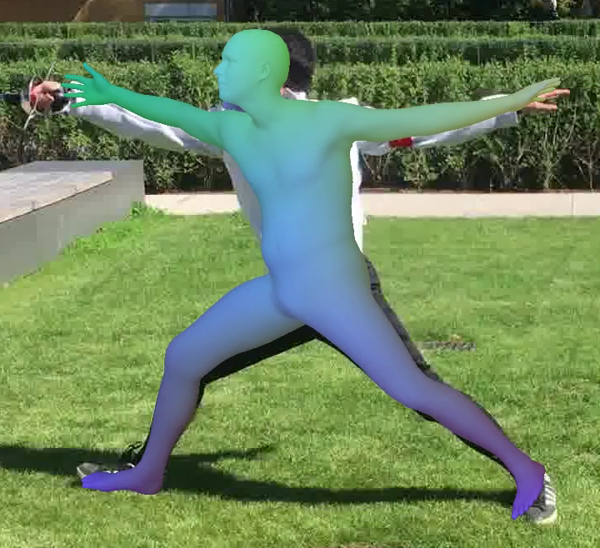} \\[2pt]
\shortstack{RGB\\\strut} & \shortstack{Shading\\\strut} & \shortstack{Canonical\\coloring} & \shortstack{Shading\\+ skeleton} & \shortstack{Shading\\+ canonical coloring} \\
\end{tabular}}
\caption{\textbf{Rendering targets in Table~\ref{tab:rendering-ablation}.} The same pose with RGB, shading, canonical coloring, a body-skeleton overlay, or blended shading and canonical colors.}
\label{fig:rendering-targets}
\end{figure}

\paragraph{Global Body Representation Transfer.} Complete geometry supervision improves recovery under occlusion (Table~\ref{tab:rendering-ablation}). Replacing RGB supervision with canonical coloring and shading reduces body/hand MPJPE from 120.0/68.4 to 106.3/57.8 mm and body PVE from 140.6 to 121.8 mm. The combined target also yields the lowest Full-input errors: 53.8/30.7 mm MPJPE and 61.3 mm PVE.

Canonical coloring and shading provide complementary cues, outperforming either alone and shading with a skeleton. Figure~\ref{fig:rendering-targets} illustrates how canonical coloring identifies surface locations, shading expresses posed geometry, and the skeleton supplies sparse joint structure. The combined target thus grounds the video representation in dense correspondence and geometry.

\begin{table}[!htb]
\centering
\caption{\textbf{Local Joint Feature Enhancement ablation.} Wan-1.3B full-window body/hand MPJPE (mm) on the ablation holdout. Local updates are absent (None), weighted by one (Ungated), or by predicted evidence scores (Evidence-Gated). Evidence supervision uses binary evidence labels (Equation~\ref{eq:evidence-supervision}).}
\label{tab:readout-gating}
\normalfont\normalsize
\setlength{\tabcolsep}{3.5pt}
\begin{tabular}{llccccc}
\toprule
\multirow{2}{*}{Decoder} & \multirow{2}{*}{Local update} & \multirow{2}{*}{Evidence supervision} & \multicolumn{2}{c}{Full} & \multicolumn{2}{c}{Occlusion} \\
\cmidrule(lr){4-5}\cmidrule(l){6-7}
& & & Body $\downarrow$ & Hands $\downarrow$ & Body $\downarrow$ & Hands $\downarrow$ \\
\midrule
Flat Query & None & No & 55.8 & 32.2 & 124.4 & 65.0 \\
Structured Query & None & No & 55.1 & 31.8 & 119.3 & 63.9 \\
Structured Query & Ungated & No & \expsecond{54.0} & 31.0 & 116.9 & 60.5 \\
Structured Query & Ungated & Yes & 54.1 & \expsecond{30.9} & 113.1 & 59.6 \\
Structured Query & Evidence-Gated & No & 54.3 & 31.2 & \expsecond{110.7} & \expsecond{59.0} \\
Structured Query & Evidence-Gated & Yes & \expbest{53.8} & \expbest{30.7} & \expbest{106.3} & \expbest{57.8} \\
\bottomrule
\end{tabular}
\end{table}

\paragraph{Structured Decoding and Local Sampling.} Table~\ref{tab:readout-gating} ablates Local Joint Feature Enhancement (Section~\ref{sec:local-enhancement}). The Structured Query Decoder replaces ordinary attention among flat task queries with part states, joint descriptors, and joint-to-part fusion (Equation~\ref{eq:state-fusion}). Combined with ungated local updates, it reduces occlusion body/hand MPJPE from 124.4/65.0 to 116.9/60.5 mm. Within this decoder, local sampling (Equation~\ref{eq:local-sampling}) reduces errors from 119.3/63.9 to 116.9/60.5 mm. Ungated updates fix the weight in Equation~\ref{eq:support-gate} to 1; the no-update variant removes this branch.

\paragraph{Evidence Gating and Supervision.} Both evidence-based weighting and evidence supervision improve local updates. With evidence supervision, gating reduces occlusion body/hand MPJPE from 113.1/59.6 to 106.3/57.8 mm. With gating, adding evidence supervision (Equation~\ref{eq:evidence-supervision}) reduces errors from 110.7/59.0 to 106.3/57.8 mm. The supervised, ungated variant trains the evidence head as an auxiliary task without using its scores for local updates; gating without evidence supervision sets $\lambda_{\rm support}=0$ (Equation~\ref{eq:readout-loss}), learning weights through reconstruction gradients. These two variants distinguish learning an evidence predictor as an auxiliary task from using its predictions to weight the local corrections. The full model combines both mechanisms, weighting local features with predicted evidence scores and supervising those scores with binary evidence labels. The full model also achieves the lowest Full-input errors, with larger gains under occlusion.

\subsection{Temporal Context under Occlusion}
\label{sec:exp-video-analysis}

\begin{table}[!htb]
\centering
\caption{\textbf{Temporal context under occlusion.} Center-frame body/hand MPJPE (mm) on the ablation holdout with full-budget Wan-1.3B. \ding{51}: original video; \ding{55}: center RGB and mask duplicated over the window, retaining the original tracks and timestamps. }
\label{tab:video-analysis}
\setlength{\tabcolsep}{4pt}
\begin{tabular}{lcccc}
\toprule
& \multicolumn{2}{c}{Body MPJPE $\downarrow$} & \multicolumn{2}{c}{Hands MPJPE $\downarrow$} \\
\cmidrule(lr){2-3}\cmidrule(l){4-5}
\rowcolor[gray]{0.92}
Temporal Context & \shortstack{\ding{51}} & \shortstack{\ding{55}}  & \shortstack{\ding{51}} & \shortstack{\ding{55}} \\
\midrule
Full observation & 56.44 & 59.70 {\textcolor{gray}{(+3.26)}} & 29.57 & 33.15 {\textcolor{gray}{(+3.58)}} \\
Occlusion & 108.92 & 121.63 {\textcolor{gray}{(+12.71)}} & 50.26 & 59.53 {\textcolor{gray}{(+9.27)}} \\
\bottomrule
\end{tabular}
\end{table}

DreamHMR uses cross-frame RGB context to improve reconstruction. Table~\ref{tab:video-analysis} evaluates the full-budget Wan-1.3B checkpoint at the same center frames after replacing every RGB frame and observation mask with the center-frame input. This intervention increases body/hand MPJPE by 12.71/9.27 mm under occlusion, compared with 3.26/3.58 mm without additional masking. The larger effect under occlusion supports the use of neighboring observations for reconstruction. This diagnostic measures the response to an inference-time change in temporal context and input distribution. The original tracks and timestamps are retained in both conditions, so the repeated sequence removes changing RGB evidence while preserving the tracking input. We score only the center frame to keep the evaluation target fixed across conditions; these errors therefore should not be compared directly with the full-window averages in the other ablations. Appendix Figure~\ref{fig:app-temporal-qualitative} shows reconstructions through successive phases of a real action.

\FloatBarrier

\section{Conclusion}
\label{sec:conclusion}

DreamHMR adapts video generation representations for complete body-and-hand recovery under occlusion. Complete human renderings transfer surface correspondence and posed geometry into the backbone, while evidence-gated local features refine contextual body-part estimates. Across four video backbones, DreamHMR improves reconstruction under occlusion and attains the lowest errors on the evaluated HMR benchmarks.

\section*{AI use statement}
\label{sec:ai-use}

In this work, we used generative AI tools to assist with method implementation, translation, and dataset cleaning and reformatting. We did not use LLMs to develop theoretical models or conceptual frameworks, propose or refine hypotheses, design or provide feedback on research methodology or experiments, support qualitative or thematic data analysis, or interpret results. Synthetic dataset generation, formulation of mathematical claims, and development or writing of mathematical proofs are not applicable to this work. Additionally, we used LLMs to create or modify scientific figures, edit software code, create research artifacts, improve manuscript readability, identify relevant literature, and search for information. The authors checked AI-assisted code and data processing for correctness, verified figures against the underlying results, and reviewed translations, language edits, and suggested references for accuracy. We take responsibility for the final content of this work, including text, claims or artifacts produced with the aid of generative AI.

\bibliography{references}
\bibliographystyle{iclr2027_conference}

\appendix
\clearpage
\section{Implementation Details}
\label{app:implementation}

\paragraph{Backbone interface.}
We use Cosmos2.5-Predict-2B (Cosmos-2B), LTX-Video-2B-0.9.6 (LTX-2B), Wan2.1-T2V-1.3B (Wan-1.3B), and Wan2.2-TI2V-5B (Wan-5B).
Each video backbone encodes the observed RGB with its frozen VAE and supplies the final Transformer features before the native velocity head. A linear projection followed by LayerNorm maps these features to the shared 512-D decoder space. Table~\ref{tab:app-backbones} summarizes the interfaces. LTX features are resampled along time by nearest-neighbor interpolation to match the 21 decoder endpoints; their spatial grid is retained. All backbones use the fixed text prompt \emph{A full-body video of a person moving naturally.}

\begin{table}[htbp]
\centering
\caption{Backbone interfaces for an 81-frame input. Endpoints refer to temporal positions before and after alignment to the shared decoder.}
\label{tab:app-backbones}
{\normalfont\normalsize
\begin{tabular}{lccc}
\toprule
Backbone & Feature dimension & Native endpoints & Decoder endpoints \\
\midrule
Wan-1.3B & 1536 & 21 & 21 \\
Wan-5B & 3072 & 21 & 21 \\
LTX-2B & 2048 & 11 & 21 \\
Cosmos-2B & 2048 & 21 & 21 \\
\bottomrule
\end{tabular}}
\end{table}

\paragraph{Structured decoder.}
The part and joint query branches each use two attention layers with eight heads, a 512-D hidden space, and a 2048-D feed-forward network. A two-layer MLP encodes box geometry, confidence, and tracking status at the corresponding endpoints. Each joint descriptor predicts a 2D location in normalized canvas coordinates. Local sampling retrieves a $3\times3$ neighborhood with offsets of one backbone feature cell, averages valid bilinear samples, and returns zero when the center or all samples are invalid. Valid samples lie within the observation region and the feature grid. The evidence head is a $1024\!\rightarrow\!256\!\rightarrow\!1$ MLP with GELU and a final sigmoid. Its score weights the local correction before joint-to-part fusion.

\paragraph{Parameter heads and temporal recovery.}
Part-specific MLPs predict 6D rotations, camera-frame pelvis position, and a presence logit at 21 endpoints. For consecutive endpoint timestamps $s_k,s_{k+1}$, the output vector $\mathbf y_k$ is interpolated at each original timestamp $s$ as
\begin{equation}
    \overline{\mathbf y}(s)=(1-\alpha)\mathbf y_k+\alpha\mathbf y_{k+1},
    \qquad \alpha=\frac{s-s_k}{s_{k+1}-s_k}.
    \label{eq:app-temporal}
\end{equation}
A residual temporal block, consisting of a kernel-3 depthwise convolution, GELU, and a pointwise convolution, refines the interpolated sequence. Its final projection is initialized to zero. The resulting 6D rotation vectors are orthonormalized into rotation matrices. A separate head predicts the shared 10D shape from part states averaged over parts and endpoints.

\paragraph{Optimization.}
The first stage trains the conditioned backbone and velocity head with the rendered-video flow-matching objective. The second stage jointly trains the backbone and structured decoder for parameter recovery while freezing the velocity head. VAE and text encoders remain frozen throughout. Table~\ref{tab:app-training} gives the settings; training samples full, static, and dynamic observations in a $2{:}1{:}1$ ratio and uses the terminal checkpoint.

\begin{table}[htbp]
\centering
\caption{Training settings. Task modules comprise the trainable conditioning projections and, during recovery, the structured decoder.}
\label{tab:app-training}
{\normalfont\normalsize
\begin{tabular}{lcc}
\toprule
Setting & Geometry adaptation & Parameter recovery \\
\midrule
Optimizer steps & 10,000 & 47,500 \\
Optimizer & AdamW & AdamW \\
Backbone / task learning rate & $10^{-6}$ / $10^{-4}$ & $10^{-6}$ / $10^{-4}$ \\
Weight decay & 0.01 & 0.01 \\
Learning-rate schedule & Constant & Constant \\
Global batch size & 24 & 24 \\
Gradient clipping & 1.0 & 1.0 \\
Hardware & $8\times$H100 80GB & $8\times$H100 80GB \\
\bottomrule
\end{tabular}}
\end{table}

\paragraph{Recovery objective.}
The parameter objective in Equation~\ref{eq:readout-loss} expands as
\begin{equation}
\begin{split}
\mathcal L_{\rm SMPLX}={}&\lambda_{\rm rot}\mathcal L_{\rm rot}
+\lambda_{\rm joint}\mathcal L_{\rm joint}
+\lambda_{\rm root}\mathcal L_{\rm root}
+\lambda_{\rm shape}\mathcal L_{\rm shape}\\
&+\lambda_{\rm 2D}\mathcal L_{\rm 2D}
+\lambda_{\rm vel}\mathcal L_{\rm vel}
+\lambda_{\rm pres}\mathcal L_{\rm pres}.
\end{split}
\label{eq:app-recovery-loss}
\end{equation}
We use hats for predictions and the superscript GT for annotations. For scalar errors $e_i$ with availability mask $m_i$, and vectors of dimension $d$, define
\begin{equation}
\operatorname{Mean}_m(e)=\frac{\sum_{i:m_i=1}e_i}{\max(1,\sum_i m_i)},
\qquad \ell_1(\mathbf u,\mathbf v)=\frac{1}{d}\|\mathbf u-\mathbf v\|_1.
\label{eq:app-loss-mean}
\end{equation}
Rotation, relative joint, localization, and relative velocity losses average within the body, left hand, and right hand before equally averaging the supervised parts:
\begin{equation}
\operatorname{PartMean}_m(e)=\frac{1}{\max(1,|\mathcal P_m|)}
\sum_{p\in\mathcal P_m}\operatorname{Mean}_{m_p}(e).
\label{eq:app-part-mean}
\end{equation}
Here $m_p$ restricts the loss-specific mask to part $p$, and $\mathcal P_m$ contains the parts with valid supervision. Counts cover the full logical batch across devices and accumulation steps; empty terms are zero. Window indices are omitted below unless needed.

\paragraph{Rotation and geometry.}
Root orientation is expressed in the camera frame and local rotations in their SMPL-X parent frames. Rotation supervision uses the geodesic angle in radians:
\begin{equation}
\begin{aligned}
 d_{\rm SO(3)}(\widehat R,R^{\rm GT})
 &=\arccos\!\left[\operatorname{clip}\!\left(
 \frac{\operatorname{tr}(\widehat R^\top R^{\rm GT})-1}{2},-1,1\right)\right],\\
 \mathcal L_{\rm rot}&=\operatorname{PartMean}_{m^{\rm rot}}\!\left(
 d_{\rm SO(3)}(\widehat R_{t,j},R^{\rm GT}_{t,j})\right).
\end{aligned}
\label{eq:app-rotation-loss}
\end{equation}
The body group includes the root and 21 body rotations; each hand contributes 15 rotations. Predicted joints $\widehat{\mathbf J}_{t,j}$ are obtained from predicted pose and shared shape through differentiable SMPL-X. Relative coordinates are
\begin{equation}
 \mathbf r_{t,j}=\mathbf J_{t,j}-\mathbf J_{t,c(j)},\qquad
 c(j)=\begin{cases}
 \text{pelvis},&j\in\text{body},\\
 \text{left wrist},&j\in\text{left hand},\\
 \text{right wrist},&j\in\text{right hand}.
 \end{cases}
\label{eq:app-relative-joints}
\end{equation}
Prediction and ground truth each use their own reference joint while retaining camera-frame axes. Relative joint and camera-frame pelvis losses use mean absolute coordinate errors in meters:
\begin{equation}
\begin{aligned}
 \mathcal L_{\rm joint}&=\operatorname{PartMean}_{m^{\rm joint}}\!\left(
 \ell_1(\widehat{\mathbf r}_{t,j},\mathbf r^{\rm GT}_{t,j})\right),\\
 \mathcal L_{\rm root}&=\operatorname{Mean}_{m^{\rm root}}\!\left(
 \ell_1(\widehat{\mathbf t}_t,\mathbf t^{\rm GT}_t)\right).
\end{aligned}
\label{eq:app-position-losses}
\end{equation}
Relative joint supervision covers 21 non-root body joints and 15 joints per hand, requiring both the joint and its reference to be valid. Shared shape contributes once per valid window $w$:
\begin{equation}
 \mathcal L_{\rm shape}=\operatorname{Mean}_{m^{\rm shape}}\!\left(
 \frac{1}{10}\|\widehat{\boldsymbol\beta}_w-\boldsymbol\beta_w^{\rm GT}\|_2^2\right).
\label{eq:app-shape-loss}
\end{equation}

\paragraph{Joint localization.}
For a canvas of width $W$ and height $H$, the ground-truth pixel coordinate $(u_{k,j},v_{k,j})$ at endpoint $k$ is normalized to canvas coordinates. The descriptor location loss is
\begin{equation}
\begin{aligned}
 \mathbf x^{\rm GT}_{k,j}&=\left(\frac{2u_{k,j}}{W-1}-1,\frac{2v_{k,j}}{H-1}-1\right),\\
 \mathcal L_{\rm 2D}&=\operatorname{PartMean}_{m^{\rm 2D}}\!\left(
 \ell_1(\widehat{\mathbf x}_{k,j},\mathbf x^{\rm GT}_{k,j})\right).
\end{aligned}
\label{eq:app-localization-loss}
\end{equation}
This term supervises all 52 descriptor locations at the 21 endpoints, with the pelvis included in the body group. Available labels outside the observation region also contribute.

\paragraph{Temporal velocity.}
For a position trajectory $\mathbf y_t$ and timestamps $s_t$ in seconds, define the adjacent-frame velocity $D_t\mathbf y=(\mathbf y_{t+1}-\mathbf y_t)/(s_{t+1}-s_t)$. The root and relative joint components use mean absolute errors in meters per second:
\begin{equation}
\begin{aligned}
 \mathcal L_{\rm vel}^{\rm root}&=\operatorname{Mean}_{m^{\rm vroot}}\!\left(
 \ell_1(D_t\widehat{\mathbf t},D_t\mathbf t^{\rm GT})\right),\\
 \mathcal L_{\rm vel}^{\rm joint}&=\operatorname{PartMean}_{m^{\rm vjoint}}\!\left(
 \ell_1(D_t\widehat{\mathbf r}_{\cdot,j},D_t\mathbf r^{\rm GT}_{\cdot,j})\right),\\
 \mathcal L_{\rm vel}&=\frac{\sum_{c\in\mathcal V}\mathcal L_{\rm vel}^{c}}{\max(1,|\mathcal V|)},
 \qquad\mathcal V\subseteq\{\mathrm{root},\mathrm{joint}\}.
\end{aligned}
\label{eq:app-velocity-loss}
\end{equation}
Here $\mathcal V$ contains the components with valid supervision. Valid pairs are consecutive source frames of the same person with increasing timestamps, positive ground-truth presence at both ends, and valid position and reference labels. Geometry and presence use all 81 frames, giving up to 80 adjacent velocity pairs.

\paragraph{Presence and evidence support.}
For logit $z$ and binary label $y$, define $\ell_{\rm BCE}(z,y)=\log(1+\exp z)-yz$. With presence logit $z_t^{\rm pres}$ and label $a_t^{\rm GT}$,
\begin{equation}
 \mathcal L_{\rm pres}=\operatorname{Mean}_{m^{\rm pres}}\!\left(
 \ell_{\rm BCE}(z_t^{\rm pres},a_t^{\rm GT})\right).
\label{eq:app-presence-loss}
\end{equation}
Support supervision balances the three groups defined in Section~\ref{sec:local-enhancement}: supported joints inside the observation region, unsupported joints inside it, and joints outside it. Let $\mathcal I_g$ contain valid endpoint--joint annotations in group $g$, $\mathcal G_+$ contain the nonempty groups, and $q_{k,j}=\sigma(z^{\rm support}_{k,j})$. Then
\begin{equation}
 \mathcal L_{\rm support}=\frac{1}{\max(1,|\mathcal G_+|)}
 \sum_{g\in\mathcal G_+}\frac{1}{|\mathcal I_g|}
 \sum_{(k,j)\in\mathcal I_g}\ell_{\rm BCE}(z^{\rm support}_{k,j},q^{\rm GT}_{k,j}).
\label{eq:app-support-loss}
\end{equation}
Presence averages valid positive and negative frames; support averages nonempty groups over the 21 endpoints. Availability depends on annotation validity and identity, independently of predicted presence or evidence scores.
\FloatBarrier
\section{Observation Construction and Evaluation}
\label{app:observations}

\paragraph{Training data.}
Table~\ref{tab:app-data} summarizes the training mixture. Frame counts refer to retained person frames before observation augmentation. We sample source sequences equally within each dataset and then select an 81-frame window of the same person. Full, static, and dynamic observations share the camera, canvas, and person scale, using a fixed $848\times480$ canvas or its portrait counterpart.

\begin{table}[htbp]
\centering
\caption{Training data and sampling proportions. Available annotations supervise the corresponding output fields.}
\label{tab:app-data}
{\normalfont\normalsize
\begin{tabular}{lrr}
\toprule
Dataset & Person frames & Sampling proportion \\
\midrule
BEDLAM & 0.82M & 18\% \\
BEDLAM2 & 1.07M & 22\% \\
EgoBody & 342K & 25\% \\
ARCTIC & 287K & 15\% \\
InterCap & 121K & 10\% \\
IDEA400 & 89K & 10\% \\
\bottomrule
\end{tabular}}
\end{table}

\begin{figure}[htbp]
\centering
{\expfloatstyle\setlength{\tabcolsep}{4pt}
\begin{tabular}{@{}ccc@{}}
\textbf{Full} & \textbf{Static} & \textbf{Dynamic} \\[3pt]
\includegraphics[width=.30\linewidth]{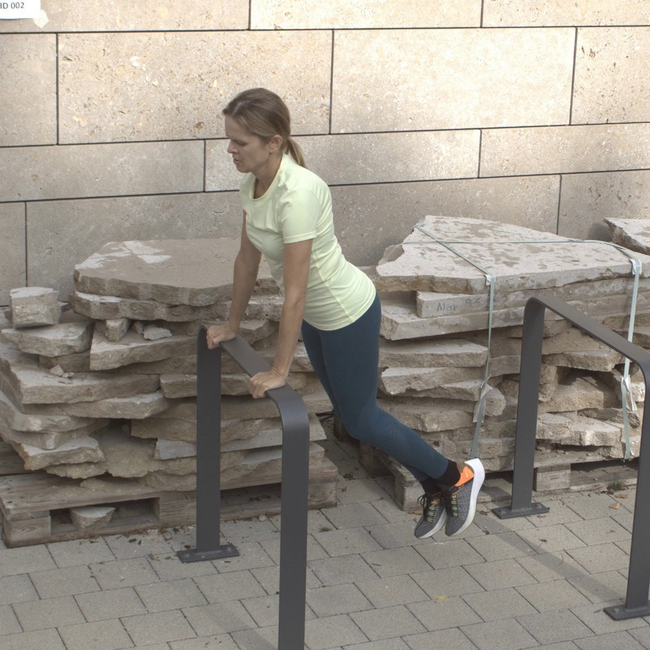} &
\includegraphics[width=.30\linewidth]{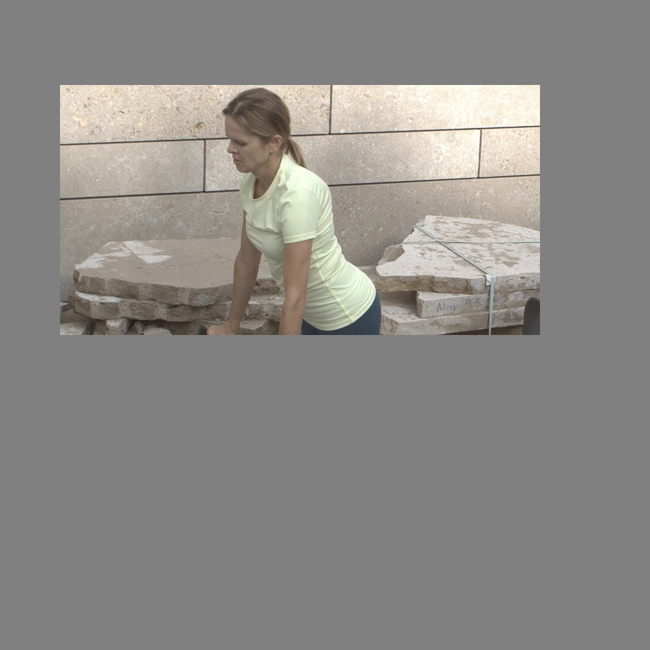} &
\includegraphics[width=.30\linewidth]{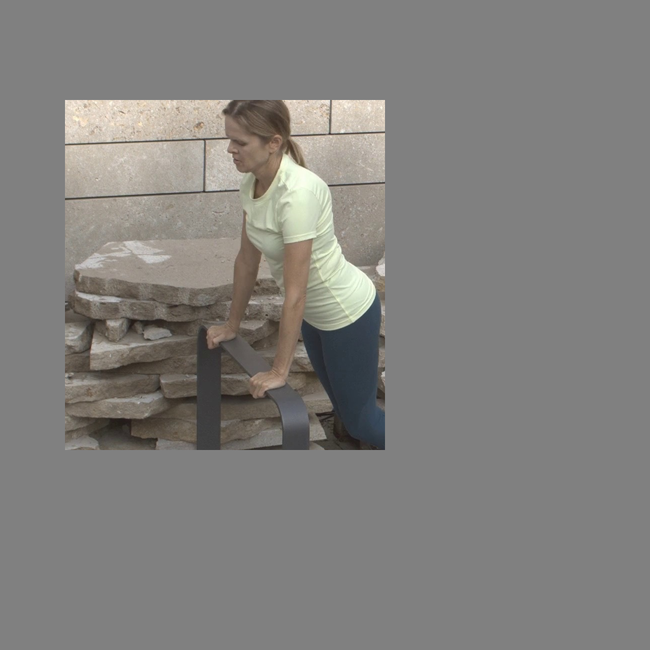} \\[4pt]
\includegraphics[width=.30\linewidth]{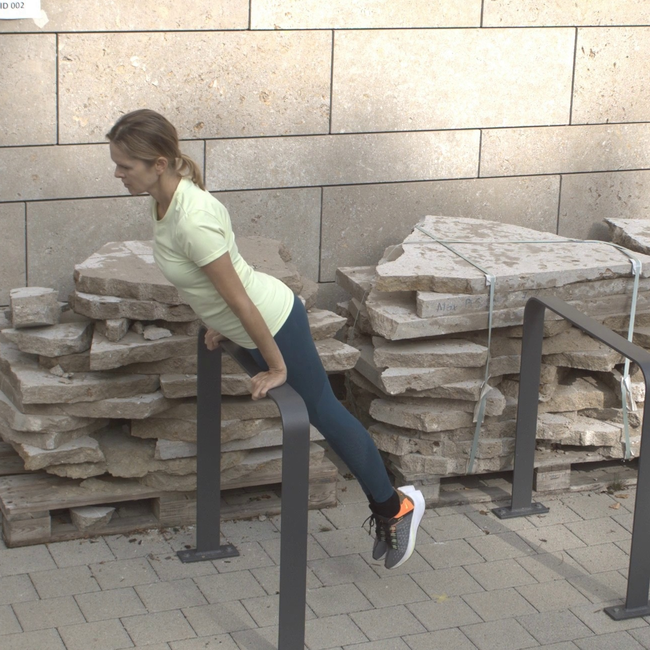} &
\includegraphics[width=.30\linewidth]{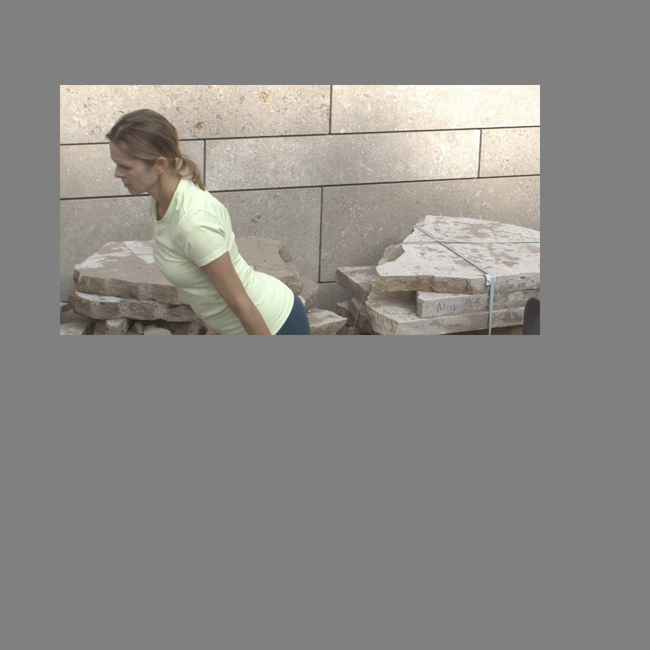} &
\includegraphics[width=.30\linewidth]{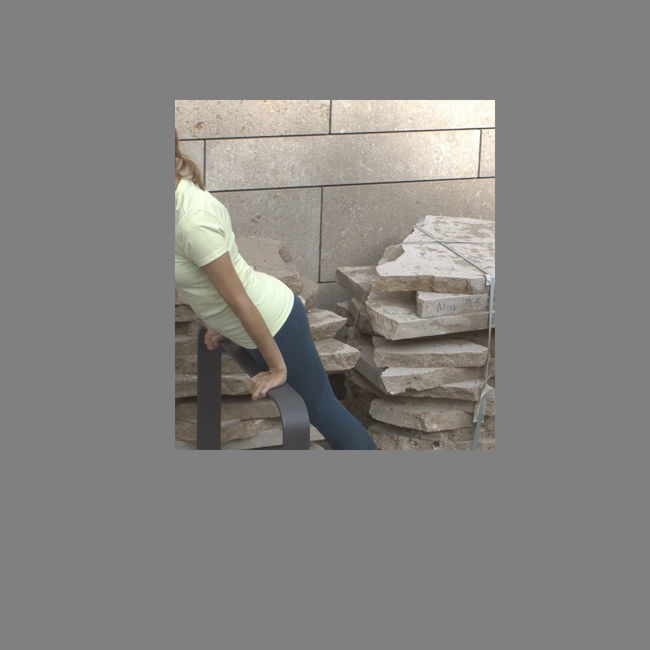} \\[3pt]
Entire canvas observed & Fixed observation region & Moving observation region
\end{tabular}}
\caption{\textbf{Observation construction.} Two RICH frames (top to bottom) under full, static, and dynamic observations. Static regions stay fixed; dynamic regions move horizontally. Pixels outside the region are neutral gray. All panels share the display crop and scale.}
\label{fig:app-observations}
\end{figure}

\paragraph{Observation regions.}
A reference box $(c_x,c_y,w,h)$ is computed once per source sequence from the median complete SMPL-X projection box over reliable frames. Eight static templates comprise two horizontal regions of size $1.25w\times0.55h$, two vertical regions of size $0.55w\times1.25h$, and four local rectangles of size $0.70w\times0.70h$. Their centers are displaced from the reference center by $0.25h$ vertically or $0.25w$ horizontally. Horizontal regions use the up/down offsets, vertical regions use left/right offsets, and local rectangles use all four directions. Eight dynamic templates move a local rectangle horizontally or vertically from these four initial positions, with a maximum displacement of $0.35w$ or $0.35h$ over the window. Regions are intersected with the physical canvas; pixels outside are set to neutral gray, $[0.5,0.5,0.5]$. Figure~\ref{fig:app-observations} illustrates the construction.

\paragraph{Tracking and supervision.}
A frozen HumanArt YOLOX-X detector with ByteTrack provides the target tracks from each observed RGB sequence. The presence label is positive when the observation region contains at least 32 visible target-person pixels. A joint receives a positive evidence label when its ground-truth projection lies in both the observation region and the target's visible instance mask. Evidence supervision uses frames with reliable, aligned visible-instance masks; other samples contribute their available geometric and 2D labels. Complete body-and-hand geometry remains supervised wherever the corresponding annotations are available.

\paragraph{Evaluation sets.}
The occlusion evaluation and ablation holdouts each contain 90 BEDLAM and 30 EgoBody source sequences. Each sequence supplies two source windows; each window yields one full, eight static, and eight dynamic observations, giving 4,080 windows per holdout. The two holdouts are disjoint from each other, training, and validation, with identities, recording sessions, source motions, and scenes separated across splits. EgoBody examples use third-person videos with body-and-hand annotations. Evaluation on widely used benchmarks uses the official 3DPW test, EMDB-1, and RICH test sets with their original RGB inputs.

\paragraph{Shared inputs and output alignment.}
All compared methods receive the same observed RGB and target tracks. Video methods process complete windows; image methods process per-frame crops from the shared tracks. We use released inference models without test-time flipping or additional temporal smoothing. SMPLest-X uses the official H40 weights, and 3DB-DINOv3 includes its released hand decoder. Public body evaluation uses 14/24/24 joints for 3DPW/EMDB-1/RICH, respectively, with common joint regressors and frozen topology mappings for SMPL-X, MHR, and SMPL-X/MANO predictions. MPJPE and PVE use pelvis alignment; PA-MPJPE additionally fits a similarity transform. Under occlusion, body MPJPE uses 21 non-root joints relative to the pelvis; hand MPJPE uses 15 joints per hand relative to the wrist, with equal left/right weighting.

\paragraph{Sequence prediction and aggregation.}
For long videos, DreamHMR centers its 81-frame windows at frames $0,40,80,\ldots$ and the final frame, replicating boundary frames where necessary. Each real frame takes the prediction from its nearest window center, with ties assigned to the earlier center. Public benchmark results average sequences equally. For occlusion evaluation, errors are averaged within each source sequence, then equally across sequences within each dataset; BEDLAM and EgoBody contribute 75\% and 25\%, respectively, and static/dynamic conditions contribute equally. The temporal-context diagnostic evaluates only the center frame: repeated inputs duplicate its RGB and observation mask while retaining the original tracks and timestamps.
\FloatBarrier
\section{Rendering Details}
\label{app:rendering}

This appendix expands the renderer summarized by Equation~\ref{eq:geometry-target}. For every annotated frame, it converts SMPL-X pose and shape parameters into one RGB image of a complete, camera-aligned human. The same procedure is applied to all frames to form the supervision video $I^{\rm rend}_{1:T}$. Canonical coloring assigns fixed template-coordinate colors to the posed mesh, and shading determines brightness from its current normals. Rasterization resolves visibility and evaluates their product at covered pixels; compositing with the source RGB produces the target frame. All rendering choices are fixed before training and shared across frames, subjects, and supervision variants.

\paragraph{Notation and camera-aligned mesh.}
The frame index is $t\in\{1,\ldots,T\}$ and the SMPL-X mesh has $N$ vertices with a fixed vertex ordering and triangle topology. We write $\boldsymbol\phi_t$ for the root orientation, $\boldsymbol\theta_t$ for the local body-and-hand pose, and $\boldsymbol\beta$ for the shape. Let $\operatorname{SMPLX}(\boldsymbol\phi,\boldsymbol\theta,\boldsymbol\beta;\mathbf s)$ return the mesh produced by SMPL-X when its native translation is $\mathbf s$, and let $\mathbf J_{0,t}$ be the pelvis returned by the same pose and shape with native translation set to zero. The annotated pelvis position in the camera frame is $\mathbf t_t$. Passing $\mathbf t_t-\mathbf J_{0,t}$ as the native translation therefore places the pelvis at $\mathbf t_t$:
\begin{equation}
    \mathbf V_t=\bigl[\mathbf V_{t,1},\ldots,\mathbf V_{t,N}\bigr]
    =\operatorname{SMPLX}\!\left(\boldsymbol\phi_t,\boldsymbol\theta_t,\boldsymbol\beta;\,
    \mathbf t_t-\mathbf J_{0,t}\right).
    \label{eq:app-mesh}
\end{equation}
Here $\mathbf V_{t,i}\in\mathbb R^3$ is the camera-frame position of vertex $i$ at frame $t$. The camera frame is centered at the optical center, with $x$ pointing right, $y$ pointing down, and $z$ pointing forward; positions are measured in meters.

The rendered mesh and the source RGB frame use the same resized-and-padded canvas. If the source intrinsics are $(f_x,f_y,c_x,c_y)$, a resize by $s$ and a pixel offset $(o_x,o_y)$ give canvas intrinsics
\begin{equation}
    \bar f_x=sf_x,\qquad \bar f_y=sf_y,\qquad
    \bar c_x=sc_x+o_x,\qquad \bar c_y=sc_y+o_y.
    \label{eq:app-intrinsics}
\end{equation}
A camera-frame point $\mathbf X=(X,Y,Z)$ with $Z>0$ projects to
\begin{equation}
    \pi_t(\mathbf X)=\left(\bar f_x X/Z+\bar c_x,\;\bar f_y Y/Z+\bar c_y\right).
    \label{eq:app-projection}
\end{equation}
The mesh and RGB input share the same canvas and camera calibration.

\paragraph{Canonical surface map.}
The canonical surface map assigns every template vertex an RGB value that encodes its canonical surface coordinates. Let $\mathbf T_i\in\mathbb R^3$ be vertex $i$ of one fixed neutral SMPL-X template in its canonical pose. Define the componentwise template extrema
\[
    \mathbf T_{\min}=\min_i\mathbf T_i,\qquad
    \mathbf T_{\max}=\max_i\mathbf T_i,
\]
where the minimum and maximum are taken independently along the template $x$, $y$, and $z$ axes. We encode these three canonical surface coordinates in the RGB channels:
\begin{equation}
    \mathbf c_i^{\rm can}
    =\frac{\mathbf T_i-\mathbf T_{\min}}
    {\mathbf T_{\max}-\mathbf T_{\min}}
    \in[0,1]^3.
    \label{eq:app-canonical-coord}
\end{equation}
The denominator and all $\mathbf c_i^{\rm can}$ values are computed once from the fixed template. Every posed mesh keeps the template's vertex ordering and topology, so vertex $i$ inherits the same canonical surface coordinates as pose and shape change its 3D position. The map is therefore shared across frames and identifies the same surface location in every pose.

\paragraph{Rasterization and visibility.}
For each covered pixel $\mathbf u$, the renderer computes screen-space barycentric weights $\lambda_{t,k}(\mathbf u)$ for the projected triangles and selects the face with minimum interpolated camera depth, with $z_{t,i}=[\mathbf V_{t,i}]_z$,
\begin{equation}
    d_{t,f}(\mathbf u)=\sum_{k=0}^{2}\lambda_{t,k}(\mathbf u)z_{t,i_k}.
    \label{eq:app-depth}
\end{equation}
Let $\Omega_t$ denote mesh coverage and $(i_0,i_1,i_2)$ the selected face. The pixel's canonical color is
\begin{equation}
    \mathbf c_t(\mathbf u)=\sum_{k=0}^{2}\lambda^\star_{t,k}(\mathbf u)\mathbf c^{\rm can}_{i_k},
    \qquad \mathbf u\in\Omega_t.
    \label{eq:app-rasterize}
\end{equation}
Coverage is evaluated over the full canvas independently of the observation mask $M_t$.

\paragraph{Pose-dependent shading.}
The canonical surface map identifies a visible surface location, while shading describes its current orientation. The normal is recomputed from the posed mesh at every frame; for face $f$ it is the normalized cross product of the two current edge vectors, and $\hat{\mathbf n}_t(\mathbf u)$ denotes the visible-point normal. A fixed camera-frame light has direction $\hat{\mathbf l}\propto(0,-0.4,-1)$, and the scalar Lambertian factor is
\begin{equation}
    \ell_t(\mathbf u)=a+d\max\!\left(0,
    \hat{\mathbf n}_t(\mathbf u)\cdot\hat{\mathbf l}\right),
    \qquad a=0.4,\quad d=0.6.
    \label{eq:app-shading}
\end{equation}
The max term removes back-facing light. Multiplying the rasterized canonical surface map by this scalar gives the shaded mesh RGB:
\begin{equation}
    \mathbf m_t(\mathbf u)=\ell_t(\mathbf u)\,\mathbf c_t(\mathbf u),
    \qquad \mathbf u\in\Omega_t.
    \label{eq:app-shaded-mesh}
\end{equation}
The light is fixed in the camera frame, whereas $\hat{\mathbf n}_t(\mathbf u)$ follows the posed geometry; brightness changes with pose while the canonical surface coordinates stay tied to the surface location.

\paragraph{Compositing the complete target.}
Let $I_t^{\rm src}(\mathbf u)$ be the source RGB frame after the full, static, or dynamic observation construction. The rendered frame replaces source pixels only where the posed mesh is visible:
\begin{equation}
    I_t^{\rm rend}(\mathbf u)=
    \begin{cases}
        \mathbf m_t(\mathbf u), & \mathbf u\in\Omega_t,\\
        I_t^{\rm src}(\mathbf u), & \mathbf u\notin\Omega_t.
    \end{cases}
    \label{eq:app-composite}
\end{equation}
Mesh coverage is independent of $M_t$, so projected human pixels outside the observed region remain in the target whenever they lie on the physical canvas. Ordinary visibility and self-occlusion still apply. Repeating Equation~\ref{eq:app-composite} for $t=1,\ldots,T$ produces $I^{\rm rend}_{1:T}$.

\paragraph{Supervision variants.}
The final target rasterizes the canonical surface coordinates from Equation~\ref{eq:app-canonical-coord} and applies the shading in Equation~\ref{eq:app-shading}. The canonical coloring variant uses $\mathbf c_t(\mathbf u)$ without the shading multiplier. The shading variant replaces every $\mathbf c_i^{\rm can}$ by the fixed color $(0.32,0.70,0.84)$ before shading. The shading + skeleton variant uses that same shaded mesh and overlays the projected 52-joint body-and-hand skeleton with fixed colors, line widths, and drawing settings. All variants share the same SMPL-X geometry, camera, canvas, source background, visibility resolution, and rasterization procedure.
\FloatBarrier
\section{Additional Results}
\label{app:additional-results}

\subsection{Effect of Backbone Initialization}
\label{sec:app-initialization}
\begin{table}[!htbp]
\centering
\caption{\textbf{Backbone initialization.} Body/hand MPJPE (mm) over all 81 frames on the ablation holdout after 10,000 task-training steps. Paired Wan runs vary DiT initialization; DINOv3-H+ is the image self-supervised reference.}
\label{tab:initialization}
\normalfont\normalsize
\setlength{\tabcolsep}{4pt}
\renewcommand{\arraystretch}{1.14}
\begin{tabular*}{\linewidth}{@{\extracolsep{\fill}}llrrrr@{}}
\toprule
& & \multicolumn{2}{c}{Full} & \multicolumn{2}{c}{Occlusion} \\
\cmidrule(lr){3-4}\cmidrule(l){5-6}
Backbone & Initialization & Body $\downarrow$ & Hands $\downarrow$ & Body $\downarrow$ & Hands $\downarrow$ \\
\midrule
DINOv3-H+ & Image SSL & 53.92 & 33.40 & 116.47 & 65.36 \\
Wan-1.3B & Random & 56.44 & 34.20 & 126.34 & 70.13 \\
Wan-1.3B & Video pretrained & 52.96 & 31.21 & 112.81 & 61.85 \\
Wan-5B & Random & 55.78 & 33.65 & 124.22 & 68.46 \\
Wan-5B & Video pretrained & 52.11 & 30.68 & 108.87 & 55.49 \\
\bottomrule
\end{tabular*}
\end{table}

Video generation pretraining provides a useful initialization for reconstruction under a fixed task-training budget. Table~\ref{tab:initialization} compares pretrained and randomly initialized DiTs within the same Wan architecture after 10,000 task-training steps, with all 81 frames evaluated. The paired runs share task data, the structured readout, evidence gating, and optimization settings; the corresponding pretrained VAE and text encoder are retained in both runs. Pretrained initialization reduces body/hand MPJPE under occlusion by 13.53/8.28 mm at 1.3B and 15.35/12.97 mm at 5B.

DINOv3-H+~\citep{simeoni2025dinov3} provides a complementary image self-supervised representation reference under the same task-training settings. The pretrained Wan-1.3B configuration reduces body/hand MPJPE under occlusion by 3.66/3.51 mm relative to this reference. This comparison evaluates the two representation choices, which differ in architecture, pretraining, and temporal encoding.
\FloatBarrier

\subsection{Evidence Gating with Static and Dynamic Observations}
\label{app:observation-dynamics}
\begin{table}[htbp]
\centering
\caption{\textbf{Evidence gating across observation patterns.} Wan-1.3B body/hand MPJPE (mm) on the ablation holdout. Ungated uses unit local weights; Gated uses supervised evidence scores. $\Delta$ is Gated minus Ungated, computed from unrounded means.}
\label{tab:app-dynamics}
{\normalfont\normalsize\setlength{\tabcolsep}{6pt}
\begin{tabular}{lrrrrrr}
\toprule
& \multicolumn{3}{c}{Body} & \multicolumn{3}{c}{Hands} \\
\cmidrule(lr){2-4}\cmidrule(lr){5-7}
Observation & Ungated & Gated & $\Delta$ & Ungated & Gated & $\Delta$ \\
\midrule
Static & 115.7 & \expbest{103.5} & $-12.2$ & 57.7 & \expbest{56.1} & $-1.6$ \\
Dynamic & 118.1 & \expbest{109.1} & $-9.0$ & 63.3 & \expbest{59.5} & $-3.8$ \\
\bottomrule
\end{tabular}}
\end{table}

Evidence gating improves reconstruction under both observation patterns (Table~\ref{tab:app-dynamics}). On the Wan-1.3B ablation holdout, it reduces body MPJPE by 12.2 mm with static regions and 9.0 mm with dynamic regions. Hand MPJPE improves by 1.6 and 3.8 mm, respectively. These results complement the pooled ablations by showing the benefit of weighting local updates when the observation region is fixed or moves across the person.

\FloatBarrier
\subsection{Additional Qualitative Comparisons}
\label{app:qualitative}

\begin{figure}[!htbp]
\centering
{\expfloatstyle
\setlength{\tabcolsep}{.7pt}
\begin{tabular}{@{}rccccccc@{}}
 & 0.00 s & 0.33 s & 0.67 s & 1.00 s & 1.33 s & 1.67 s & 2.00 s \\[2pt]
\raisebox{17pt}{RGB} & \includegraphics[width=.119\linewidth]{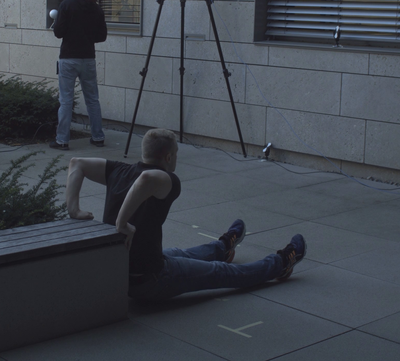} & \includegraphics[width=.119\linewidth]{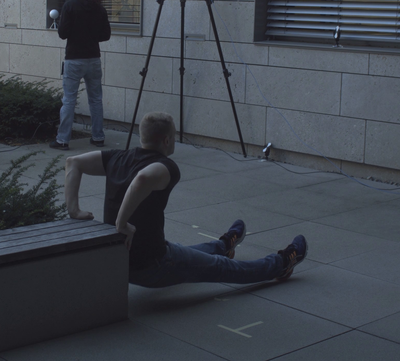} & \includegraphics[width=.119\linewidth]{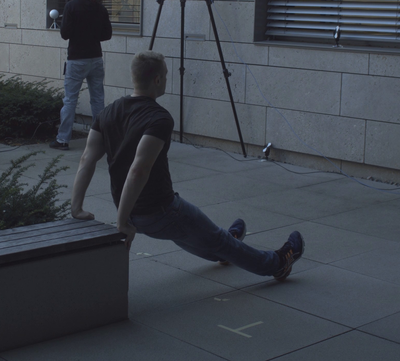} & \includegraphics[width=.119\linewidth]{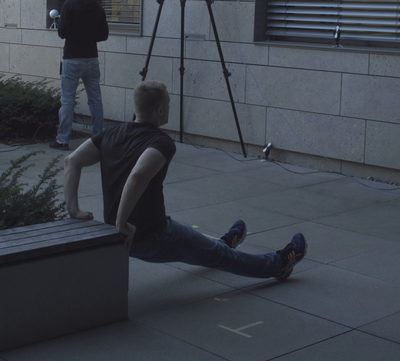} & \includegraphics[width=.119\linewidth]{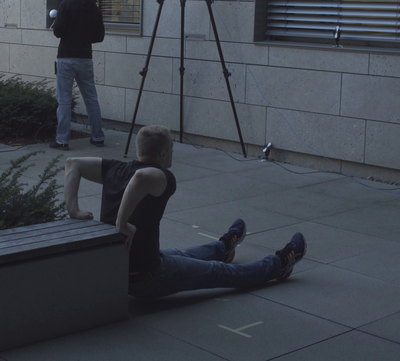} & \includegraphics[width=.119\linewidth]{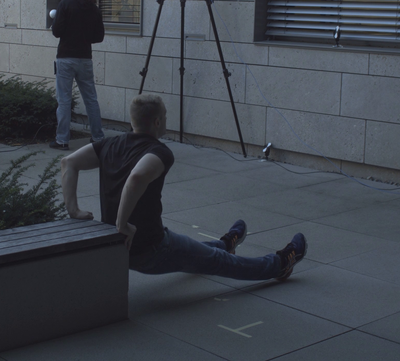} & \includegraphics[width=.119\linewidth]{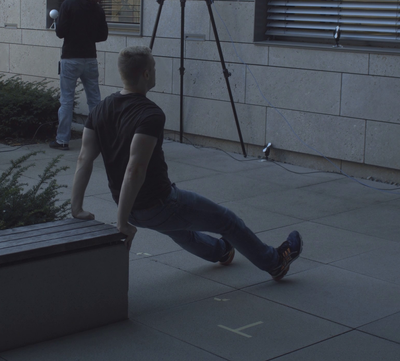} \\[1pt]
\raisebox{17pt}{\shortstack[r]{Hand4-\\Whole++}} & \includegraphics[width=.119\linewidth]{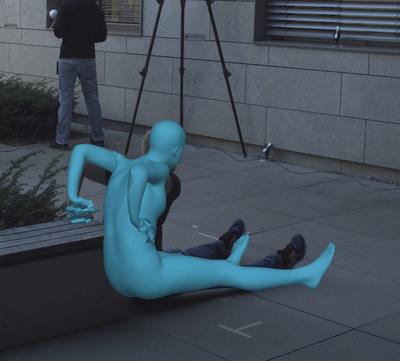} & \includegraphics[width=.119\linewidth]{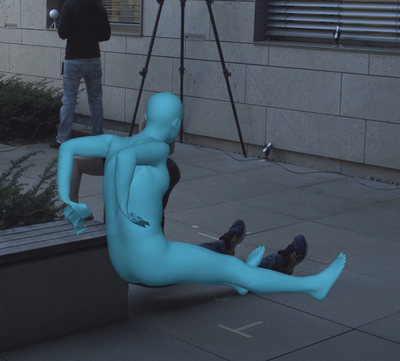} & \includegraphics[width=.119\linewidth]{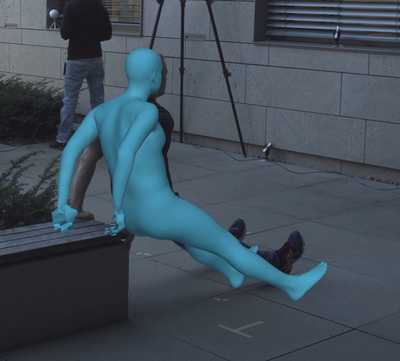} & \includegraphics[width=.119\linewidth]{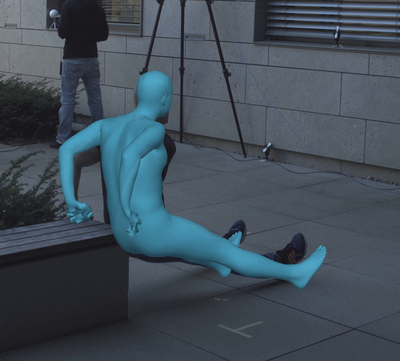} & \includegraphics[width=.119\linewidth]{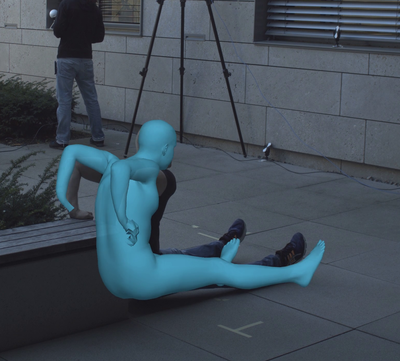} & \includegraphics[width=.119\linewidth]{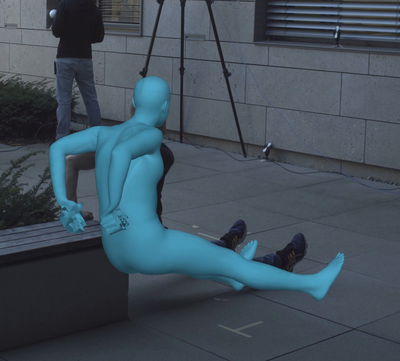} & \includegraphics[width=.119\linewidth]{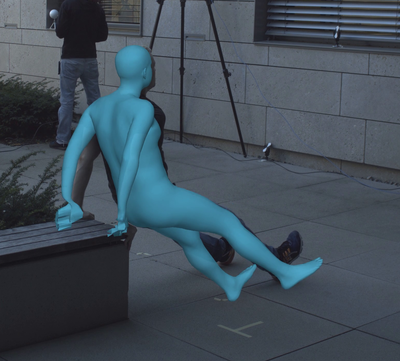} \\[1pt]
\raisebox{17pt}{SMPLest-X} & \includegraphics[width=.119\linewidth]{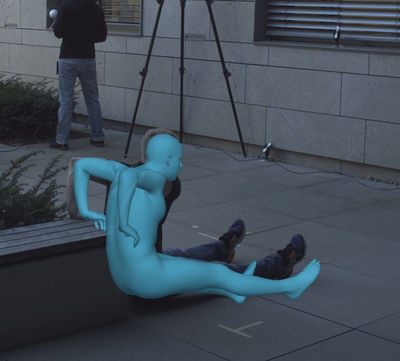} & \includegraphics[width=.119\linewidth]{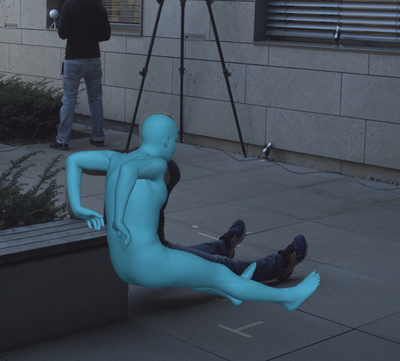} & \includegraphics[width=.119\linewidth]{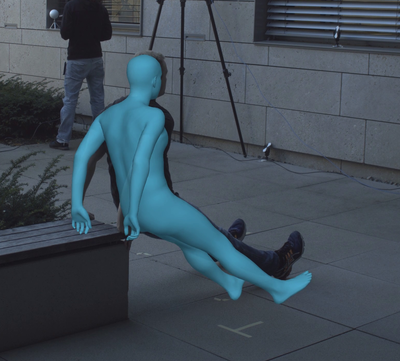} & \includegraphics[width=.119\linewidth]{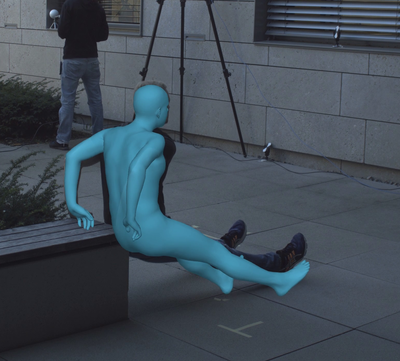} & \includegraphics[width=.119\linewidth]{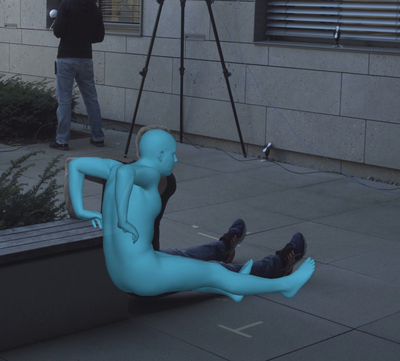} & \includegraphics[width=.119\linewidth]{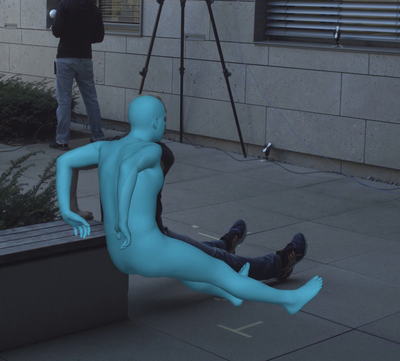} & \includegraphics[width=.119\linewidth]{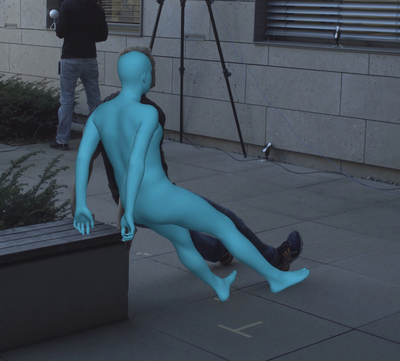} \\[1pt]
\raisebox{17pt}{DreamHMR} & \includegraphics[width=.119\linewidth]{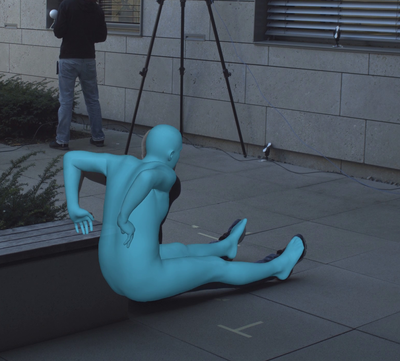} & \includegraphics[width=.119\linewidth]{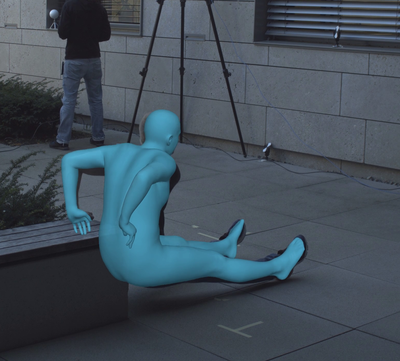} & \includegraphics[width=.119\linewidth]{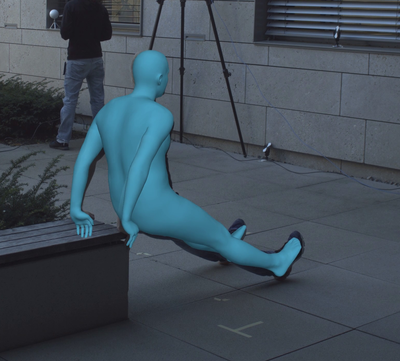} & \includegraphics[width=.119\linewidth]{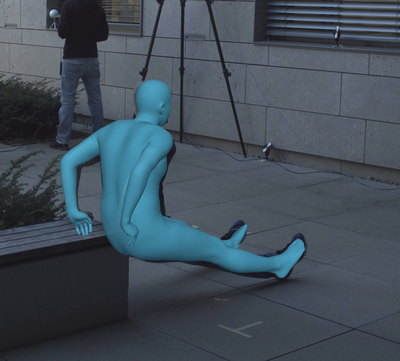} & \includegraphics[width=.119\linewidth]{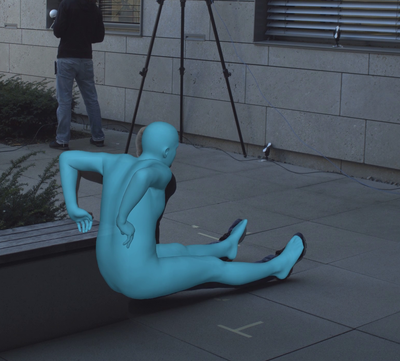} & \includegraphics[width=.119\linewidth]{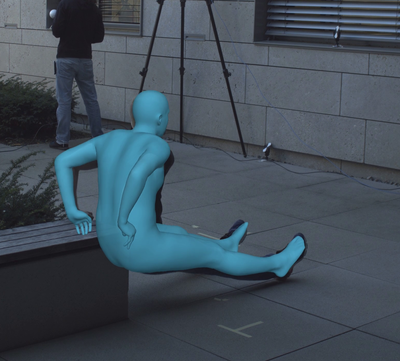} & \includegraphics[width=.119\linewidth]{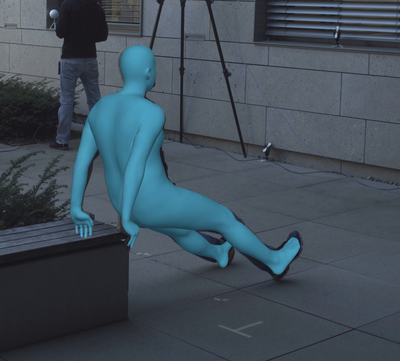} \\[1pt]
\raisebox{17pt}{GT} & \includegraphics[width=.119\linewidth]{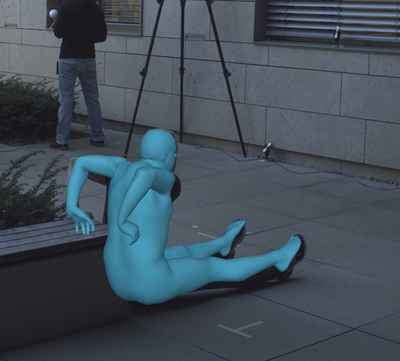} & \includegraphics[width=.119\linewidth]{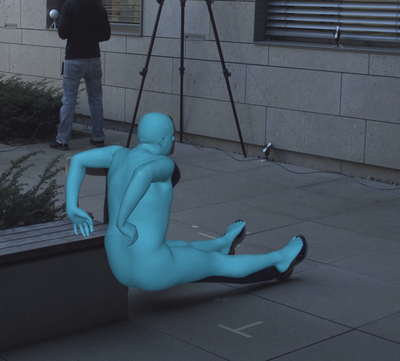} & \includegraphics[width=.119\linewidth]{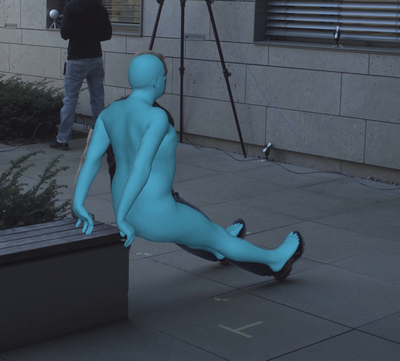} & \includegraphics[width=.119\linewidth]{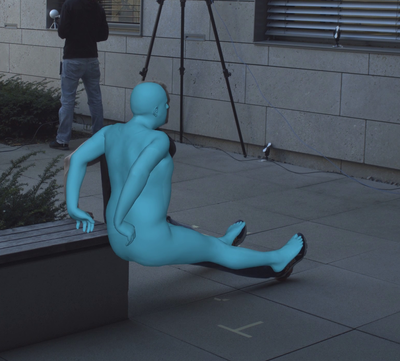} & \includegraphics[width=.119\linewidth]{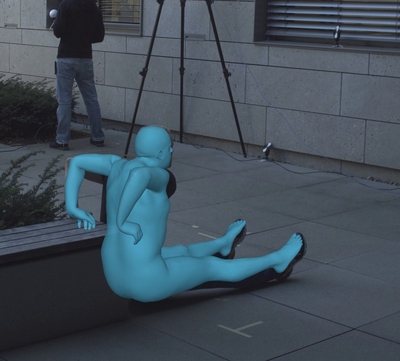} & \includegraphics[width=.119\linewidth]{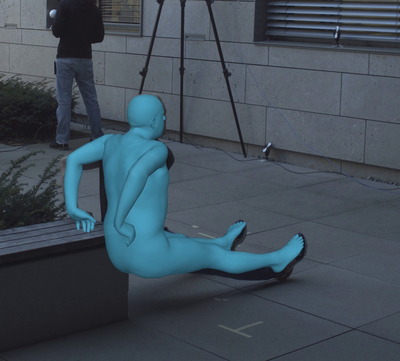} & \includegraphics[width=.119\linewidth]{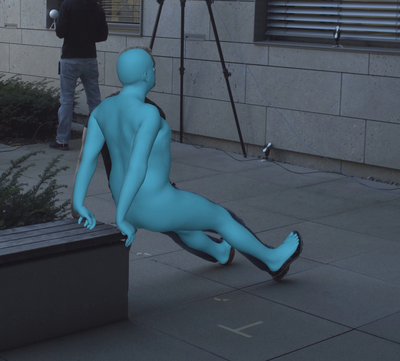} \\[1pt]
\end{tabular}}
\caption{\textbf{Reconstruction over time.} Wan-1.3B on a RICH bench-dip sequence: seven frames at 0.33-second intervals span two seconds. All methods share a fixed crop and scale throughout the sequence. GT denotes ground-truth meshes.}
\label{fig:app-temporal-qualitative}
\end{figure}

Figure~\ref{fig:app-temporal-qualitative} follows a two-second bench-dip motion. DreamHMR keeps the hands gripping the bench edge and both feet inside the shoes, while both baselines leave the shoes uncovered.

\FloatBarrier

\begin{figure}[p]
\centering
{\expfont\fontsize{8}{9}\selectfont
\setlength{\tabcolsep}{2pt}
\renewcommand{\arraystretch}{0.9}
\begin{tabular}{@{}cccc@{}}
\textbf{(a)} RGB & Hand4Whole++ & SMPLest-X & GT \\[1pt]
\includegraphics[width=.238\linewidth,height=.107\textheight,keepaspectratio]{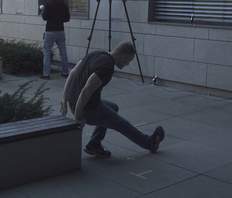} & \includegraphics[width=.238\linewidth,height=.107\textheight,keepaspectratio]{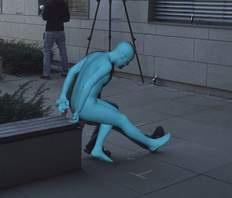} & \includegraphics[width=.238\linewidth,height=.107\textheight,keepaspectratio]{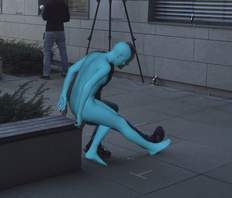} & \includegraphics[width=.238\linewidth,height=.107\textheight,keepaspectratio]{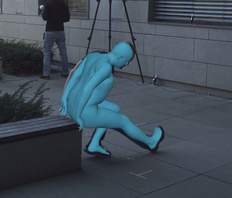} \\[1pt]
\textcolor{expteal}{\textbf{Wan-1.3B}} & \textcolor{expteal}{\textbf{Wan-5B}} & \textcolor{expteal}{\textbf{LTX-2B}} & \textcolor{expteal}{\textbf{Cosmos-2B}} \\[1pt]
\includegraphics[width=.238\linewidth,height=.107\textheight,keepaspectratio]{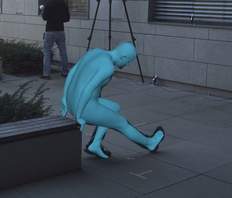} & \includegraphics[width=.238\linewidth,height=.107\textheight,keepaspectratio]{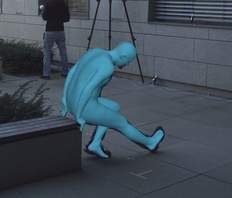} & \includegraphics[width=.238\linewidth,height=.107\textheight,keepaspectratio]{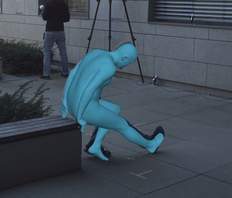} & \includegraphics[width=.238\linewidth,height=.107\textheight,keepaspectratio]{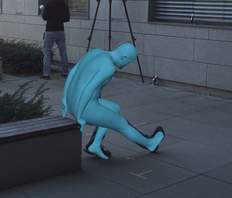} \\[1pt]
\end{tabular}\par\vspace{1pt}
\begin{tabular}{@{}cccc@{}}
\llap{\textbf{(b)}\hspace{4pt}}\includegraphics[width=.238\linewidth,height=.107\textheight,keepaspectratio]{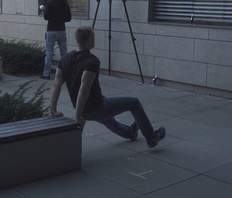} & \includegraphics[width=.238\linewidth,height=.107\textheight,keepaspectratio]{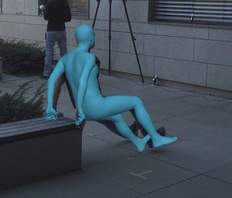} & \includegraphics[width=.238\linewidth,height=.107\textheight,keepaspectratio]{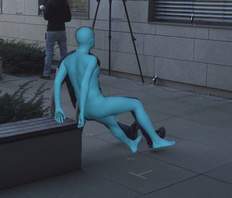} & \includegraphics[width=.238\linewidth,height=.107\textheight,keepaspectratio]{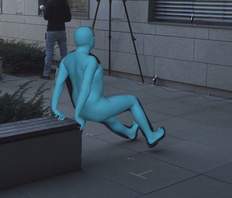} \\[1pt]
\includegraphics[width=.238\linewidth,height=.107\textheight,keepaspectratio]{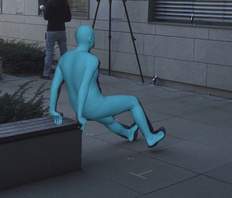} & \includegraphics[width=.238\linewidth,height=.107\textheight,keepaspectratio]{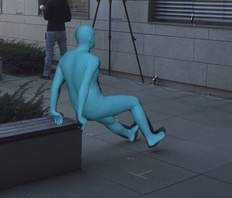} & \includegraphics[width=.238\linewidth,height=.107\textheight,keepaspectratio]{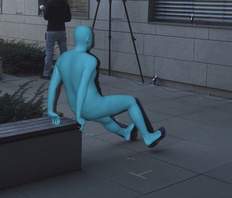} & \includegraphics[width=.238\linewidth,height=.107\textheight,keepaspectratio]{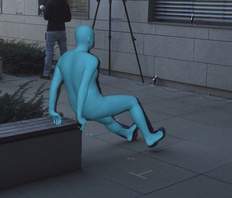} \\[1pt]
\end{tabular}\par\vspace{1pt}
\begin{tabular}{@{}cccc@{}}
\llap{\textbf{(c)}\hspace{4pt}}\includegraphics[width=.238\linewidth,height=.107\textheight,keepaspectratio]{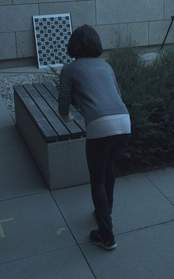} & \includegraphics[width=.238\linewidth,height=.107\textheight,keepaspectratio]{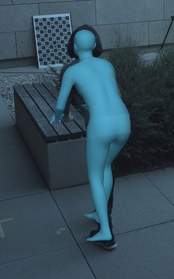} & \includegraphics[width=.238\linewidth,height=.107\textheight,keepaspectratio]{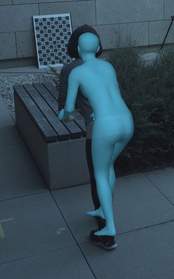} & \includegraphics[width=.238\linewidth,height=.107\textheight,keepaspectratio]{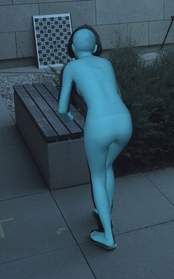} \\[1pt]
\includegraphics[width=.238\linewidth,height=.107\textheight,keepaspectratio]{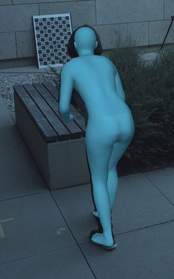} & \includegraphics[width=.238\linewidth,height=.107\textheight,keepaspectratio]{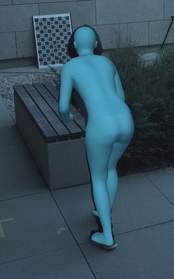} & \includegraphics[width=.238\linewidth,height=.107\textheight,keepaspectratio]{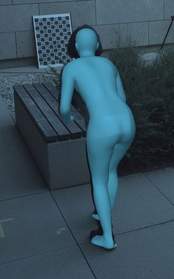} & \includegraphics[width=.238\linewidth,height=.107\textheight,keepaspectratio]{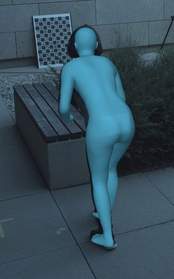} \\[1pt]
\end{tabular}\par\vspace{1pt}
\begin{tabular}{@{}cccc@{}}
\llap{\textbf{(d)}\hspace{4pt}}\includegraphics[width=.238\linewidth,height=.107\textheight,keepaspectratio]{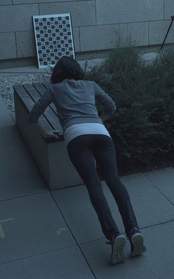} & \includegraphics[width=.238\linewidth,height=.107\textheight,keepaspectratio]{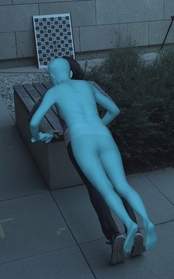} & \includegraphics[width=.238\linewidth,height=.107\textheight,keepaspectratio]{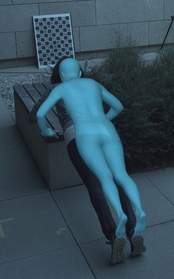} & \includegraphics[width=.238\linewidth,height=.107\textheight,keepaspectratio]{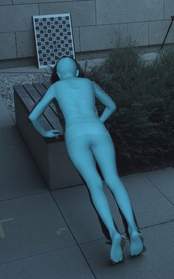} \\[1pt]
\includegraphics[width=.238\linewidth,height=.107\textheight,keepaspectratio]{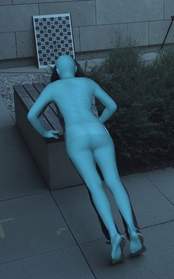} & \includegraphics[width=.238\linewidth,height=.107\textheight,keepaspectratio]{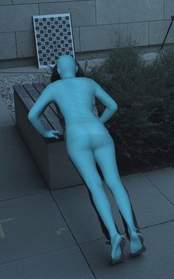} & \includegraphics[width=.238\linewidth,height=.107\textheight,keepaspectratio]{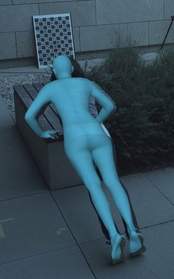} & \includegraphics[width=.238\linewidth,height=.107\textheight,keepaspectratio]{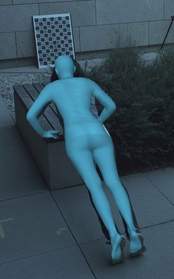} \\[1pt]
\end{tabular}\par\vspace{1pt}
}
\caption{\textbf{Qualitative comparisons across scenes.} DreamHMR with four video backbones, compared with Hand4Whole++ and SMPLest-X. RGB and ground truth are shown for reference.}
\label{fig:app-four-dips}
\end{figure}

Figure~\ref{fig:app-four-dips} compares all four DreamHMR configurations with Hand4Whole++ and SMPLest-X across eight videos, including bench support, walking, reaching, and foreground occlusions.

\begin{figure}[p]
\centering
{\expfont\fontsize{8}{9}\selectfont
\setlength{\tabcolsep}{2pt}
\renewcommand{\arraystretch}{0.9}
\textbf{Figure~\ref{fig:app-four-dips} (continued)}\par\smallskip
\begin{tabular}{@{}cccc@{}}
\textbf{(e)} RGB & Hand4Whole++ & SMPLest-X & GT \\[1pt]
\includegraphics[width=.238\linewidth,height=.107\textheight,keepaspectratio]{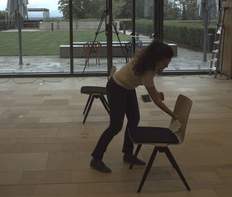} & \includegraphics[width=.238\linewidth,height=.107\textheight,keepaspectratio]{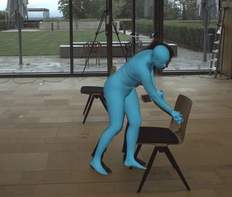} & \includegraphics[width=.238\linewidth,height=.107\textheight,keepaspectratio]{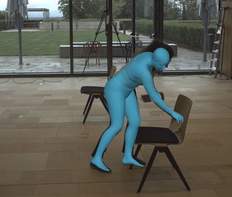} & \includegraphics[width=.238\linewidth,height=.107\textheight,keepaspectratio]{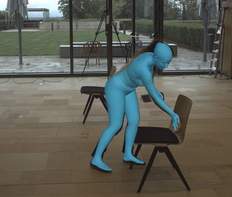} \\[1pt]
\textcolor{expteal}{\textbf{Wan-1.3B}} & \textcolor{expteal}{\textbf{Wan-5B}} & \textcolor{expteal}{\textbf{LTX-2B}} & \textcolor{expteal}{\textbf{Cosmos-2B}} \\[1pt]
\includegraphics[width=.238\linewidth,height=.107\textheight,keepaspectratio]{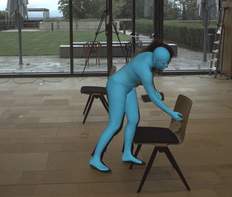} & \includegraphics[width=.238\linewidth,height=.107\textheight,keepaspectratio]{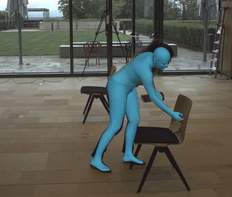} & \includegraphics[width=.238\linewidth,height=.107\textheight,keepaspectratio]{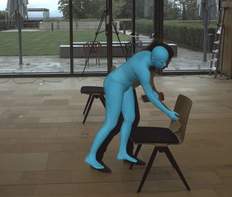} & \includegraphics[width=.238\linewidth,height=.107\textheight,keepaspectratio]{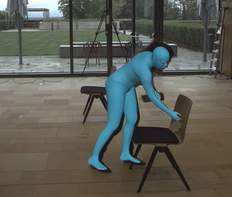} \\[1pt]
\end{tabular}\par\vspace{1pt}
\begin{tabular}{@{}cccc@{}}
\llap{\textbf{(f)}\hspace{4pt}}\includegraphics[width=.238\linewidth,height=.107\textheight,keepaspectratio]{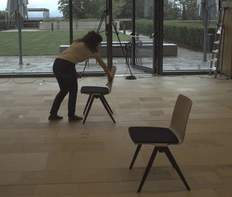} & \includegraphics[width=.238\linewidth,height=.107\textheight,keepaspectratio]{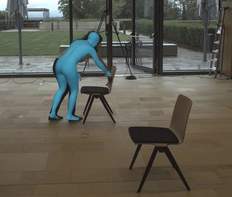} & \includegraphics[width=.238\linewidth,height=.107\textheight,keepaspectratio]{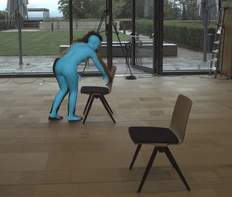} & \includegraphics[width=.238\linewidth,height=.107\textheight,keepaspectratio]{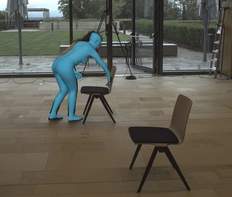} \\[1pt]
\includegraphics[width=.238\linewidth,height=.107\textheight,keepaspectratio]{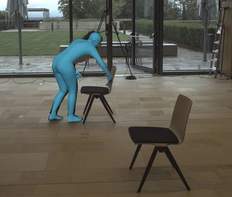} & \includegraphics[width=.238\linewidth,height=.107\textheight,keepaspectratio]{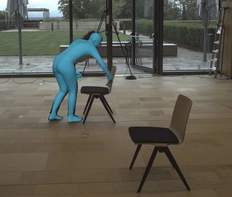} & \includegraphics[width=.238\linewidth,height=.107\textheight,keepaspectratio]{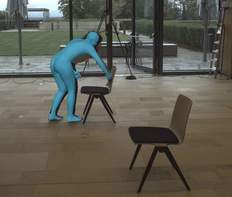} & \includegraphics[width=.238\linewidth,height=.107\textheight,keepaspectratio]{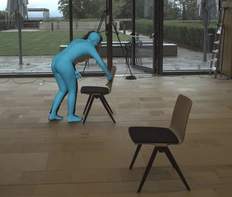} \\[1pt]
\end{tabular}\par\vspace{1pt}
\begin{tabular}{@{}cccc@{}}
\llap{\textbf{(g)}\hspace{4pt}}\includegraphics[width=.238\linewidth,height=.107\textheight,keepaspectratio]{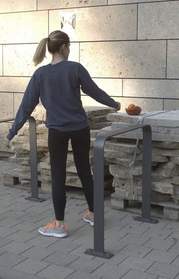} & \includegraphics[width=.238\linewidth,height=.107\textheight,keepaspectratio]{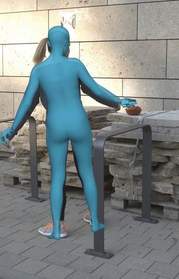} & \includegraphics[width=.238\linewidth,height=.107\textheight,keepaspectratio]{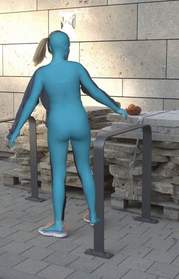} & \includegraphics[width=.238\linewidth,height=.107\textheight,keepaspectratio]{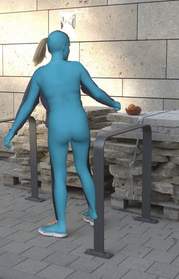} \\[1pt]
\includegraphics[width=.238\linewidth,height=.107\textheight,keepaspectratio]{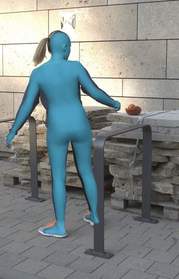} & \includegraphics[width=.238\linewidth,height=.107\textheight,keepaspectratio]{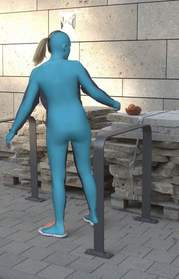} & \includegraphics[width=.238\linewidth,height=.107\textheight,keepaspectratio]{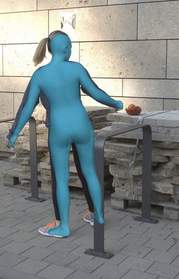} & \includegraphics[width=.238\linewidth,height=.107\textheight,keepaspectratio]{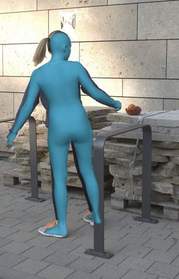} \\[1pt]
\end{tabular}\par\vspace{1pt}
\begin{tabular}{@{}cccc@{}}
\llap{\textbf{(h)}\hspace{4pt}}\includegraphics[width=.238\linewidth,height=.107\textheight,keepaspectratio]{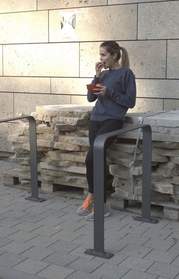} & \includegraphics[width=.238\linewidth,height=.107\textheight,keepaspectratio]{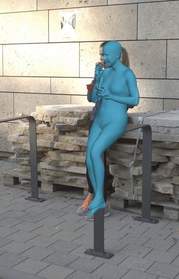} & \includegraphics[width=.238\linewidth,height=.107\textheight,keepaspectratio]{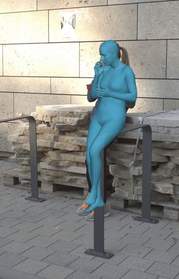} & \includegraphics[width=.238\linewidth,height=.107\textheight,keepaspectratio]{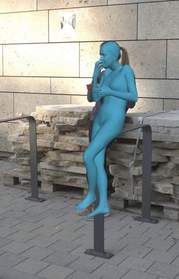} \\[1pt]
\includegraphics[width=.238\linewidth,height=.107\textheight,keepaspectratio]{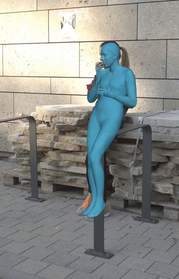} & \includegraphics[width=.238\linewidth,height=.107\textheight,keepaspectratio]{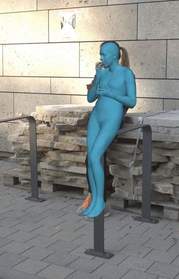} & \includegraphics[width=.238\linewidth,height=.107\textheight,keepaspectratio]{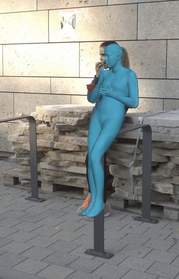} & \includegraphics[width=.238\linewidth,height=.107\textheight,keepaspectratio]{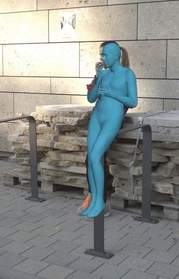} \\[1pt]
\end{tabular}\par\vspace{1pt}
}
\end{figure}
\begin{figure}[p]
\centering
{\expfont\fontsize{8}{9}\selectfont
\setlength{\tabcolsep}{2pt}
\renewcommand{\arraystretch}{0.9}
\textbf{Figure~\ref{fig:app-four-dips} (continued)}\par\smallskip
\begin{tabular}{@{}cccc@{}}
\textbf{(i)} RGB & Hand4Whole++ & SMPLest-X & GT \\[1pt]
\includegraphics[width=.238\linewidth,height=.107\textheight,keepaspectratio]{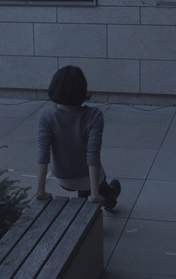} & \includegraphics[width=.238\linewidth,height=.107\textheight,keepaspectratio]{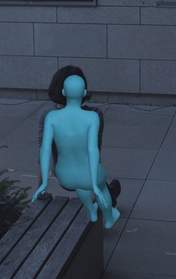} & \includegraphics[width=.238\linewidth,height=.107\textheight,keepaspectratio]{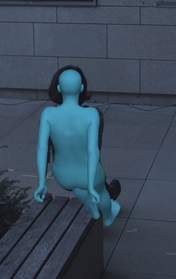} & \includegraphics[width=.238\linewidth,height=.107\textheight,keepaspectratio]{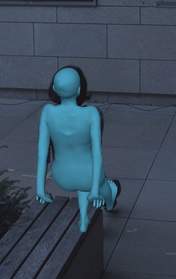} \\[1pt]
\textcolor{expteal}{\textbf{Wan-1.3B}} & \textcolor{expteal}{\textbf{Wan-5B}} & \textcolor{expteal}{\textbf{LTX-2B}} & \textcolor{expteal}{\textbf{Cosmos-2B}} \\[1pt]
\includegraphics[width=.238\linewidth,height=.107\textheight,keepaspectratio]{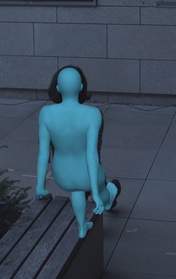} & \includegraphics[width=.238\linewidth,height=.107\textheight,keepaspectratio]{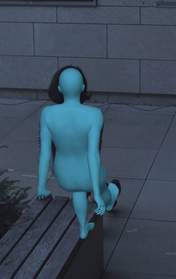} & \includegraphics[width=.238\linewidth,height=.107\textheight,keepaspectratio]{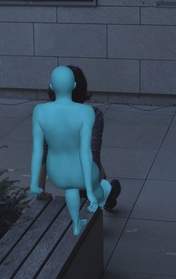} & \includegraphics[width=.238\linewidth,height=.107\textheight,keepaspectratio]{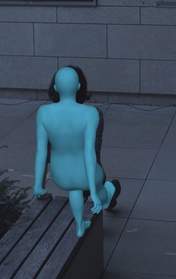} \\[1pt]
\end{tabular}\par\vspace{1pt}
\begin{tabular}{@{}cccc@{}}
\llap{\textbf{(j)}\hspace{4pt}}\includegraphics[width=.238\linewidth,height=.107\textheight,keepaspectratio]{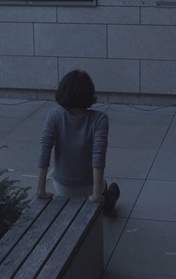} & \includegraphics[width=.238\linewidth,height=.107\textheight,keepaspectratio]{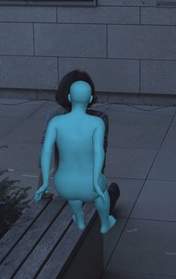} & \includegraphics[width=.238\linewidth,height=.107\textheight,keepaspectratio]{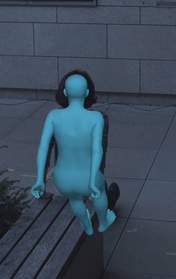} & \includegraphics[width=.238\linewidth,height=.107\textheight,keepaspectratio]{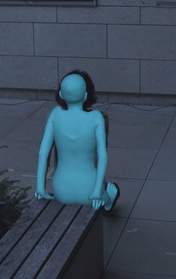} \\[1pt]
\includegraphics[width=.238\linewidth,height=.107\textheight,keepaspectratio]{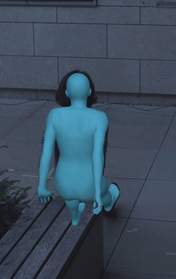} & \includegraphics[width=.238\linewidth,height=.107\textheight,keepaspectratio]{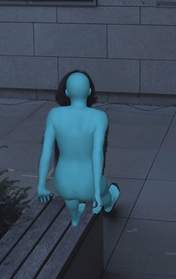} & \includegraphics[width=.238\linewidth,height=.107\textheight,keepaspectratio]{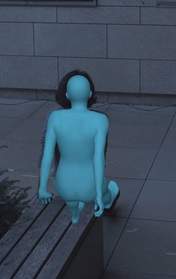} & \includegraphics[width=.238\linewidth,height=.107\textheight,keepaspectratio]{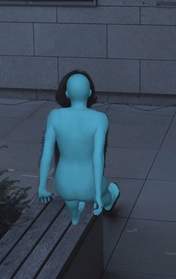} \\[1pt]
\end{tabular}\par\vspace{1pt}
\begin{tabular}{@{}cccc@{}}
\llap{\textbf{(k)}\hspace{4pt}}\includegraphics[width=.238\linewidth,height=.107\textheight,keepaspectratio]{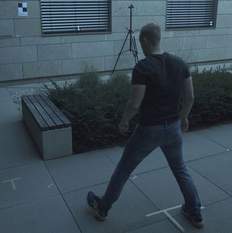} & \includegraphics[width=.238\linewidth,height=.107\textheight,keepaspectratio]{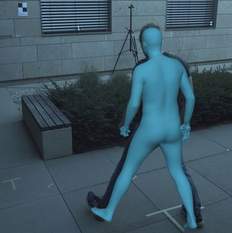} & \includegraphics[width=.238\linewidth,height=.107\textheight,keepaspectratio]{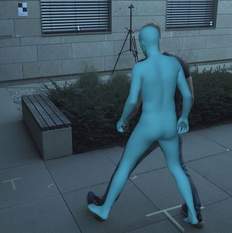} & \includegraphics[width=.238\linewidth,height=.107\textheight,keepaspectratio]{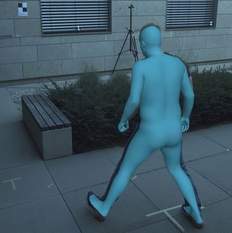} \\[1pt]
\includegraphics[width=.238\linewidth,height=.107\textheight,keepaspectratio]{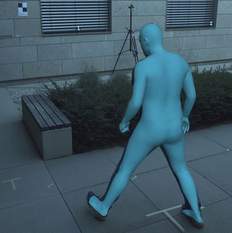} & \includegraphics[width=.238\linewidth,height=.107\textheight,keepaspectratio]{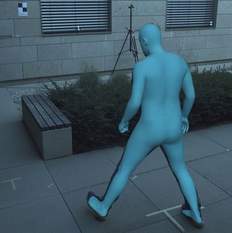} & \includegraphics[width=.238\linewidth,height=.107\textheight,keepaspectratio]{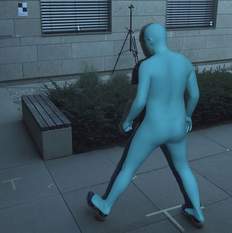} & \includegraphics[width=.238\linewidth,height=.107\textheight,keepaspectratio]{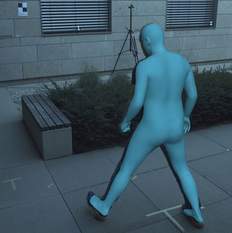} \\[1pt]
\end{tabular}\par\vspace{1pt}
\begin{tabular}{@{}cccc@{}}
\llap{\textbf{(l)}\hspace{4pt}}\includegraphics[width=.238\linewidth,height=.107\textheight,keepaspectratio]{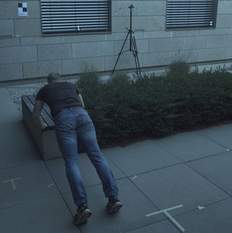} & \includegraphics[width=.238\linewidth,height=.107\textheight,keepaspectratio]{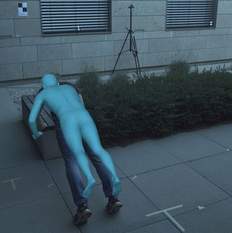} & \includegraphics[width=.238\linewidth,height=.107\textheight,keepaspectratio]{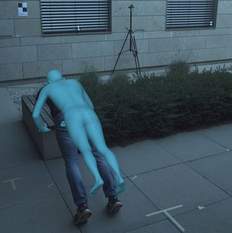} & \includegraphics[width=.238\linewidth,height=.107\textheight,keepaspectratio]{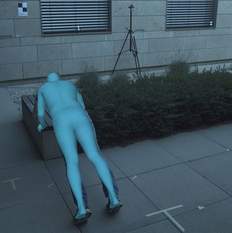} \\[1pt]
\includegraphics[width=.238\linewidth,height=.107\textheight,keepaspectratio]{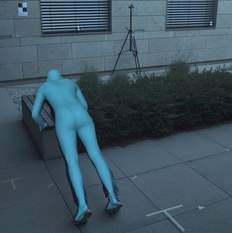} & \includegraphics[width=.238\linewidth,height=.107\textheight,keepaspectratio]{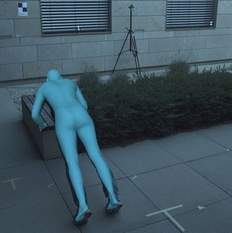} & \includegraphics[width=.238\linewidth,height=.107\textheight,keepaspectratio]{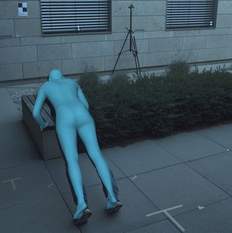} & \includegraphics[width=.238\linewidth,height=.107\textheight,keepaspectratio]{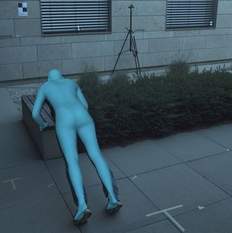} \\[1pt]
\end{tabular}\par\vspace{1pt}
}
\end{figure}
\begin{figure}[p]
\centering
{\expfont\fontsize{8}{9}\selectfont
\setlength{\tabcolsep}{2pt}
\renewcommand{\arraystretch}{0.9}
\textbf{Figure~\ref{fig:app-four-dips} (continued)}\par\smallskip
\begin{tabular}{@{}cccc@{}}
\textbf{(m)} RGB & Hand4Whole++ & SMPLest-X & GT \\[1pt]
\includegraphics[width=.238\linewidth,height=.107\textheight,keepaspectratio]{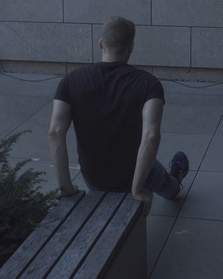} & \includegraphics[width=.238\linewidth,height=.107\textheight,keepaspectratio]{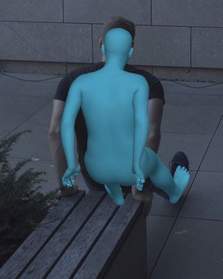} & \includegraphics[width=.238\linewidth,height=.107\textheight,keepaspectratio]{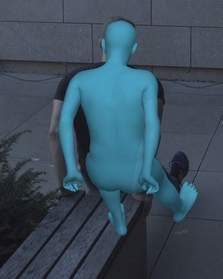} & \includegraphics[width=.238\linewidth,height=.107\textheight,keepaspectratio]{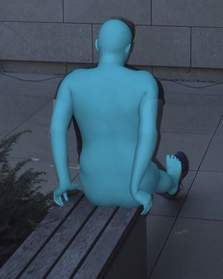} \\[1pt]
\textcolor{expteal}{\textbf{Wan-1.3B}} & \textcolor{expteal}{\textbf{Wan-5B}} & \textcolor{expteal}{\textbf{LTX-2B}} & \textcolor{expteal}{\textbf{Cosmos-2B}} \\[1pt]
\includegraphics[width=.238\linewidth,height=.107\textheight,keepaspectratio]{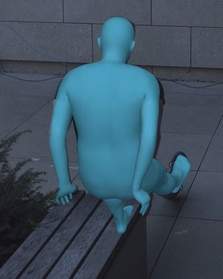} & \includegraphics[width=.238\linewidth,height=.107\textheight,keepaspectratio]{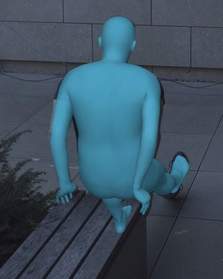} & \includegraphics[width=.238\linewidth,height=.107\textheight,keepaspectratio]{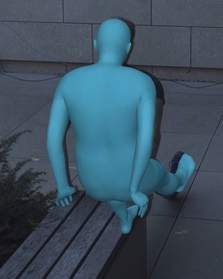} & \includegraphics[width=.238\linewidth,height=.107\textheight,keepaspectratio]{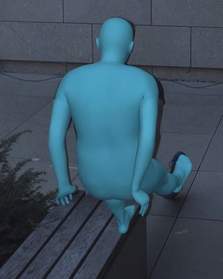} \\[1pt]
\end{tabular}\par\vspace{1pt}
\begin{tabular}{@{}cccc@{}}
\llap{\textbf{(n)}\hspace{4pt}}\includegraphics[width=.238\linewidth,height=.107\textheight,keepaspectratio]{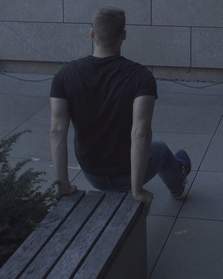} & \includegraphics[width=.238\linewidth,height=.107\textheight,keepaspectratio]{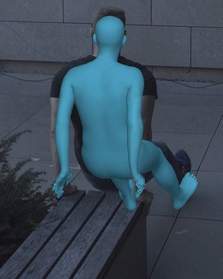} & \includegraphics[width=.238\linewidth,height=.107\textheight,keepaspectratio]{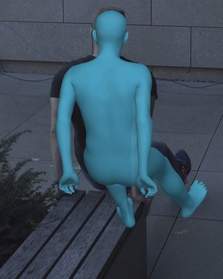} & \includegraphics[width=.238\linewidth,height=.107\textheight,keepaspectratio]{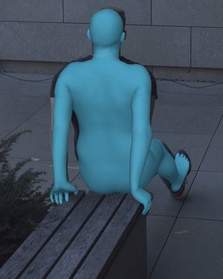} \\[1pt]
\includegraphics[width=.238\linewidth,height=.107\textheight,keepaspectratio]{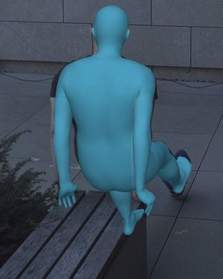} & \includegraphics[width=.238\linewidth,height=.107\textheight,keepaspectratio]{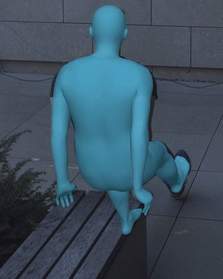} & \includegraphics[width=.238\linewidth,height=.107\textheight,keepaspectratio]{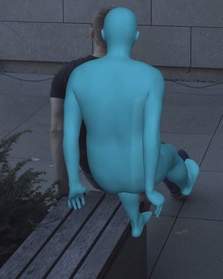} & \includegraphics[width=.238\linewidth,height=.107\textheight,keepaspectratio]{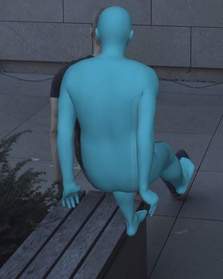} \\[1pt]
\end{tabular}\par\vspace{1pt}
\begin{tabular}{@{}cccc@{}}
\llap{\textbf{(o)}\hspace{4pt}}\includegraphics[width=.238\linewidth,height=.107\textheight,keepaspectratio]{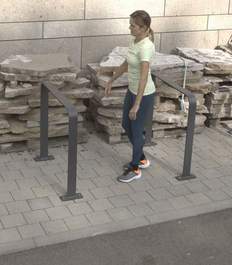} & \includegraphics[width=.238\linewidth,height=.107\textheight,keepaspectratio]{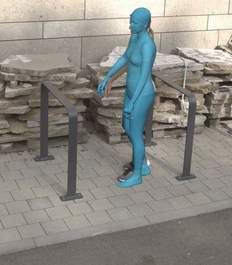} & \includegraphics[width=.238\linewidth,height=.107\textheight,keepaspectratio]{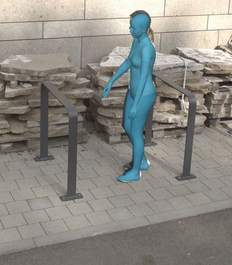} & \includegraphics[width=.238\linewidth,height=.107\textheight,keepaspectratio]{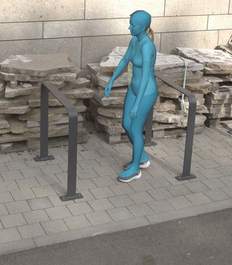} \\[1pt]
\includegraphics[width=.238\linewidth,height=.107\textheight,keepaspectratio]{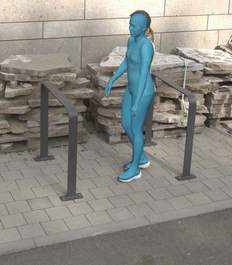} & \includegraphics[width=.238\linewidth,height=.107\textheight,keepaspectratio]{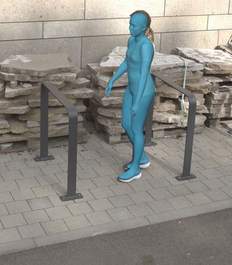} & \includegraphics[width=.238\linewidth,height=.107\textheight,keepaspectratio]{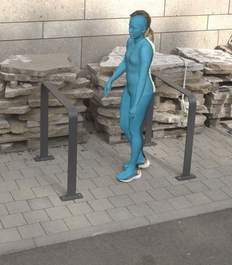} & \includegraphics[width=.238\linewidth,height=.107\textheight,keepaspectratio]{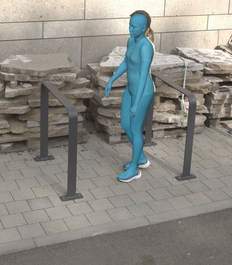} \\[1pt]
\end{tabular}\par\vspace{1pt}
\begin{tabular}{@{}cccc@{}}
\llap{\textbf{(p)}\hspace{4pt}}\includegraphics[width=.238\linewidth,height=.107\textheight,keepaspectratio]{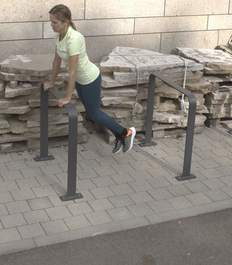} & \includegraphics[width=.238\linewidth,height=.107\textheight,keepaspectratio]{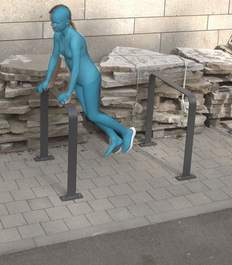} & \includegraphics[width=.238\linewidth,height=.107\textheight,keepaspectratio]{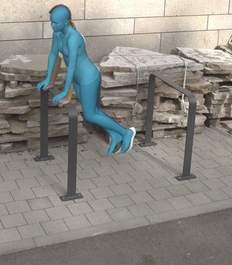} & \includegraphics[width=.238\linewidth,height=.107\textheight,keepaspectratio]{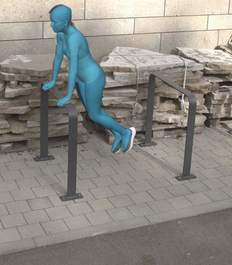} \\[1pt]
\includegraphics[width=.238\linewidth,height=.107\textheight,keepaspectratio]{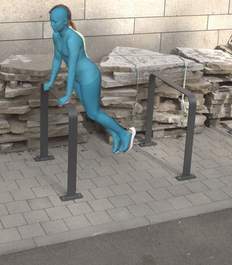} & \includegraphics[width=.238\linewidth,height=.107\textheight,keepaspectratio]{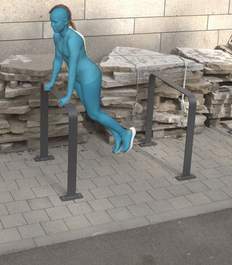} & \includegraphics[width=.238\linewidth,height=.107\textheight,keepaspectratio]{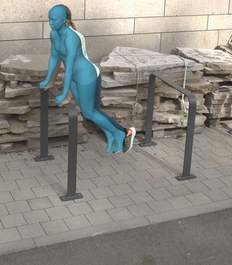} & \includegraphics[width=.238\linewidth,height=.107\textheight,keepaspectratio]{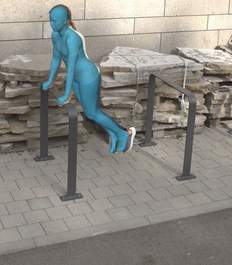} \\[1pt]
\end{tabular}\par\vspace{1pt}
}
\end{figure}

\FloatBarrier

\end{document}